\documentclass[conference]{IEEEtran}
\usepackage{amsmath,amsfonts}
\usepackage{geometry}                		
\usepackage[parfill]{parskip}    		        
\usepackage{graphicx}				
\usepackage{amssymb}				
\usepackage{amsmath}
\usepackage{textcomp}
\usepackage[style=numeric,natbib=true]{biblatex} 
\usepackage{bbm}
\usepackage{tabularx}
\usepackage{latexsym}
\usepackage{subcaption}
\usepackage{gensymb}
\usepackage{fancyhdr}
\usepackage{multicol}
\usepackage[makeroom]{cancel}
\usepackage{eucal}
\usepackage{float}
\usepackage{parskip}
\usepackage{textcomp} 
\title{Autonomous Mobile Plant Watering Robot : A Kinematic Approach}
\author{Justin London}
\usepackage{geometry}                		
\usepackage{tabularx}
\usepackage{stmaryrd}
\usepackage{wasysym} 
\usepackage{xurl}

\usepackage{microtype}
\usepackage{yfonts}
\usepackage{listings} 
\usepackage{multirow}
\usepackage{xcolor}
\usepackage{pdfpages}
\usepackage{graphicx} 
\usepackage[utf8]{inputenc}
\usepackage [english]{babel}
\usepackage [autostyle, english = american]{csquotes}
\graphicspath{{./Images/}} 
\usepackage{mathtools}
\usepackage{amsmath}
\usepackage{titlesec}
\usepackage{upgreek} 
\usepackage{amsmath} 
\usepackage{amssymb}

\PassOptionsToPackage{hyphens, spaces, obeyspaces}{url} 
\usepackage[final=true,backref=true,colorlinks=true,linkcolor=blue, urlcolor=blue, citecolor=blue, breaklinks=true]{hyperref}
\hypersetup{
    pagebackref=true,
    colorlinks=true,
    citecolor=blue,
    linkcolor=blue,
    filecolor=magenta,      
    urlcolor=blue,
}
\hypersetup{breaklinks=true}
\usepackage{breakurl}

\def\xHyphenate#1#2\wholeString {\if#1$%
    \else\transform{#1}%
    \takeTheRest#2\ofTheString\fi}

\def\takeTheRest#1\ofTheString\fi
{\fi \xHyphenate#1\wholeString}

\def\transform#1{\url{#1}\hskip 0pt plus 1pt}

\definecolor{dkgreen}{rgb}{0,0.6,0}
\definecolor{gray}{rgb}{0.5,0.5,0.5}
\definecolor{mauve}{rgb}{0.58,0,0.82}
\definecolor{lightsilver}{rgb}{0.96,0.96,0.98} 
\definecolor{blond}{rgb}{1, 0.98, 0.8}
\lstdefinestyle{R} {
backgroundcolor=\color{blond},
language=Python,
basicstyle=\tiny,
} 
\titleformat{\chapter}[display]
{\normalfont\Huge\bfseries\raggedright}{\chaptertitlename\ 3}
{15pt}{\Huge}

\title{Autonomous Mobile Plant Watering Robot: A Kinematic Approach}
\IEEEtitleabstractindextext{
\begin{abstract}
Plants need regular and the appropriate amount of watering to thrive and survive.  While agricultural robots exist that can spray water on plants and crops such as the , they are expensive and have limited mobility and/or functionality.  We introduce a novel autonomous mobile plant watering robot that uses a 6 degree of freedom (DOF) manipulator, connected to a 4 wheel drive alloy chassis, to be able to hold a garden hose, recognize and detect plants, and to water them with the appropriate amount of water by being able to insert a soil humidity/moisture sensor into the soil.  The robot uses Jetson Nano and Arduino microcontroller and real sense camera to perform computer vision to detect plants using real-time YOLOv5 with the Pl@ntNet-300K dataset.  The robot uses LIDAR for object and collision avoideance and does not need to move on a pre-defined path and can keep track of which plants it has watered.  We provide the Denavit-Hartenberg (DH) Table, forward kinematics, differential driving kinematics, and inverse kinematics along with simulation and experiment results.
\end{abstract}
}

\begin{document}

\author{\IEEEauthorblockN{Justin London \\}
\IEEEauthorblockA{\textit{Department of Electrical Engineering and Computer Science} \\
\textit{University of North Dakota}\\
Grand Forks, North Dakota USA \\
justin.london@und.edu}
}
 
\maketitle

\IEEEdisplaynontitleabstractindextext
 
\section{Introduction} 
\ \ Plants are essential to the sustenance of life.   Plants provide essential nutrients like vitamins, minerals, and fiber and provide important fruits and vegetables for nutrient-rich sustenance.  Through the process of photosynthesis, plants release oxygen into the atmosphere, absorb carbon dioxide, provide nutrients to animals, and regulate the water cycle.  In particular, plants use sunlight, water, and carbon dioxide to synthesize oxygen and energy in the form of glucose (sugar) thereby transforming light energy into chemical energy needed by all life forms.   Plants acts as natural air purifiers and without plants, animal life could not survive.  Plants have a waxy residue called a cuticle that protects plants from dehydration.  There are 380,000 plants species in the world of which the majority are vascular.  Most agriculture resolves around growing plants for human consumption.  Plants require proper watering to survive and thrive.  

As global warming causes temperatures to increase worldwide, water consumption required by plants and vegetation also increases.   To automate the process of watering plants, we introduce a robot that can autonomously recognize, detect, and water plants.  While the concept of the mobile plant watering robot has been discussed \cite{Aswani:2012} along with smart garden water systems \cite{Angelopoulos:2011}, there are no commercially available mobile watering robots in the market that can perform these important tasks except for expensive UGV agriculture robots for farmers such as the XAG R150 which costs over \$30,000.\footnote{See \url{https://www.xa.com/en/xauv_r150}}

An autonomous mobile plant water robot is introduced that overcomes the limitations for such a robot to exist.  First, real-time object detection of plants is needed via YOLOv5+ algorithms, GPUs are required.  Therefore, the objection detection is performed using machine learning (ML)/computer vision on a Jetson Nano microcontroller.  Second, the robot has two water reservoir tanks.   One reservoir tank is 18L and connected a 60V diaphragm miniature water pump and the other reservoir tank is 11L and connected to a 30V diaphragm miniature water pump.  

The 18L tank weighs 39.6 lbs when full and the 11L tank weights 24.2 lbs when full.  The 30V pump has a 100 PSI and pumps water at 0.8 GPM (3 LPM) while the 60V pump has a 116 PSI and pumps water at 1.35 GPM (5 LPM).
The robot has car-type chassis with 4 DC planetary gear motors with 4 wheel drive.  The chassis has length of 500 mm (1.64 ft) and a width of 400 mm (1.31 ft).   The chassis and gear motors are designed to handle the weight and load of the water reservoirs, 12V rechargable lead acid batteries, and other components.  When commercially designed for the market, the robot will use a custom designed single PCB board.

The robot uses an assembled AbbIrb120-type manipulator with end-effector grippers that can hold, manipulate, and angle hoses, water tubes, and a soil moisture-temperature-humidity probe sensor. Once a plant is detected, the robot must determine the type and amount of watering a plant needs.  If the plant has a large area/density such as a bush, palm, or tree, the robot uses a hose with a jet water spray nozzle. If the plant is potted, the robot will go to the plant and insert a soil moisture probe.  If the soil moisture is below a certain threshold, the robot will use a water tube that is inserted into the plant's soil.  The robot uses a relay switch to turn the water pumps on and off depending on the type of watering the plant needs (sprayed from hose or poured from tubing).   

\section{Controllers}

    The robot uses various microcontrollers and control boards:
    \begin{enumerate}
        \item Jetson Nano microcontroller - Serial Communication with Arduino for computer vision,for DC motor and sensor control,
        \item Arduino Uno microcontroller
    \end{enumerate}

    The robot uses 2 L298 H-bridge motor control drivers. The L298 boards are integrated monolithic circuit in a 15- lead Multiwatt and PowerSO20 packages. It is a high voltage, high current dual full-bridge driver designed to accept standard TTL logic levels and drive inductive loads such as relays, solenoids, DC and stepping motors.

\section{Sensors}

    The robot uses numerous sensors to gather data from its environment including:
    \begin{enumerate}
        \item 3D Orbbec Astra S Depth 3D ROS Scanning Camera : for real time sensing of environment
        \item SlamTec RPLiDAR 360 Degree 2D Lidar Sensor: for smart object detection and avoidance. 15Hz scanning frequency rate, 12 meters measuring radius (distance), 16000 measuring frequency. 
        \item Acogedor Soil Moisture Sensor and FS304-SHT3X Digital Temperature and Humidity Sensor Probe : measure soil and atmosphere 
        \end{enumerate} 

    The sensors are connected to the pins in the serial bus in the back of the Jetson Nano which in turn is controlled by input devices consisting of a wireless game controller joystick such as a PS2 or Foxglove. 

\section{Actuators and Motors}

\begin{figure}[H]
   \centering	\includegraphics[width=0.55\columnwidth]{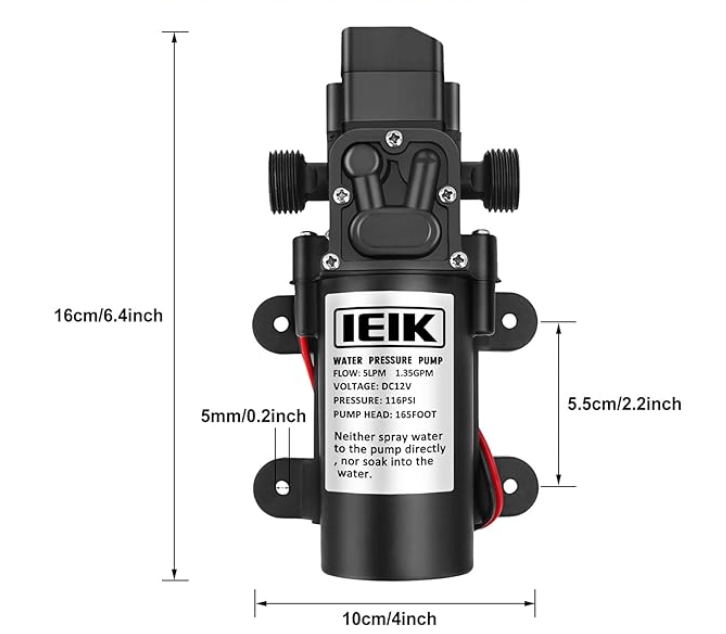} 
    \caption{Pumps}
    \label{fig:pump}
\end{figure}

\begin{figure}[H]
   \centering	\includegraphics[width=0.5\columnwidth]{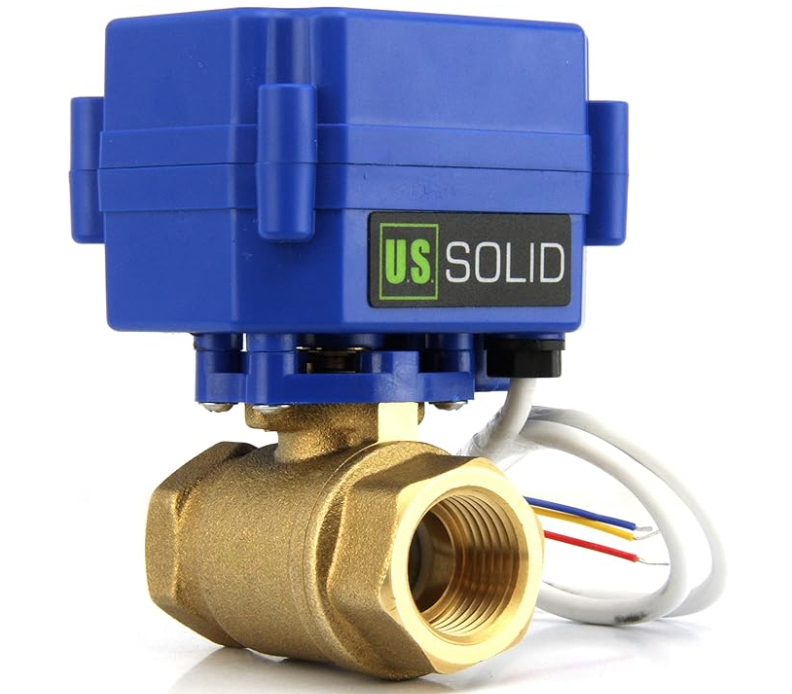} 
    \caption{Motor Ball Valve}
    \label{fig:valves}
\end{figure}

\begin{figure}[H]
   \centering	
    \includegraphics[width=0.85\columnwidth]{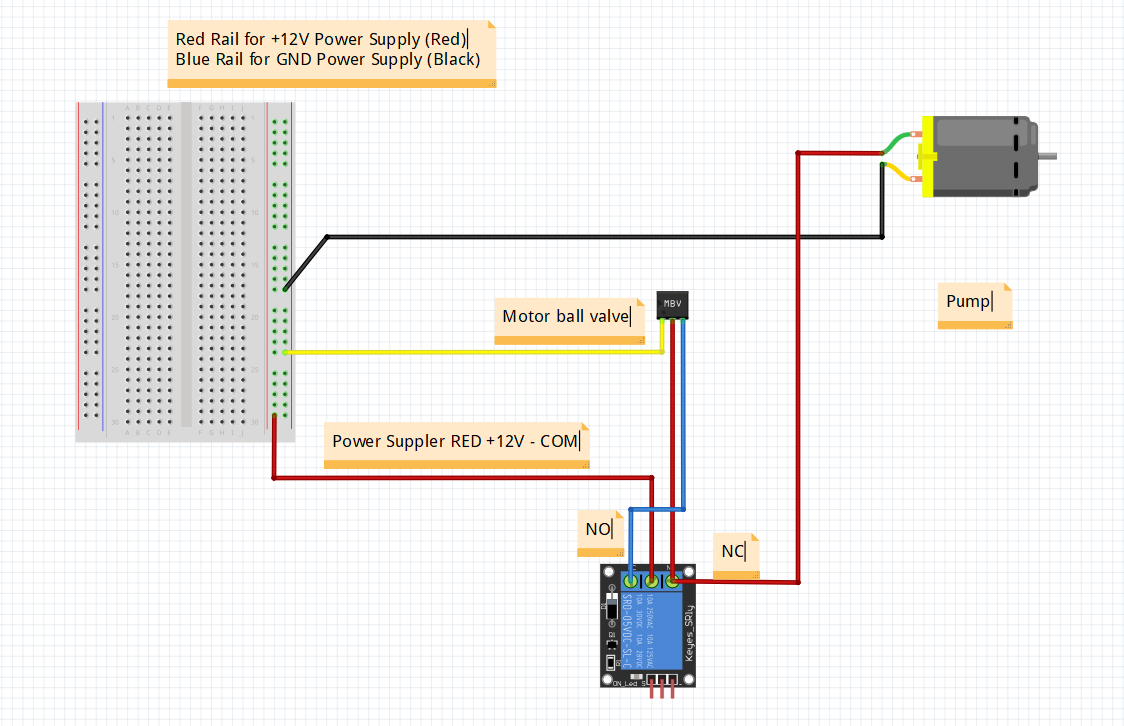} 
    \caption{Wiring Diagram of Motors Pumps and Valve to relay switch}
    \label{fig:wiring}
\end{figure}

\begin{figure}[H]
   \centering	\includegraphics[width=0.5\columnwidth]{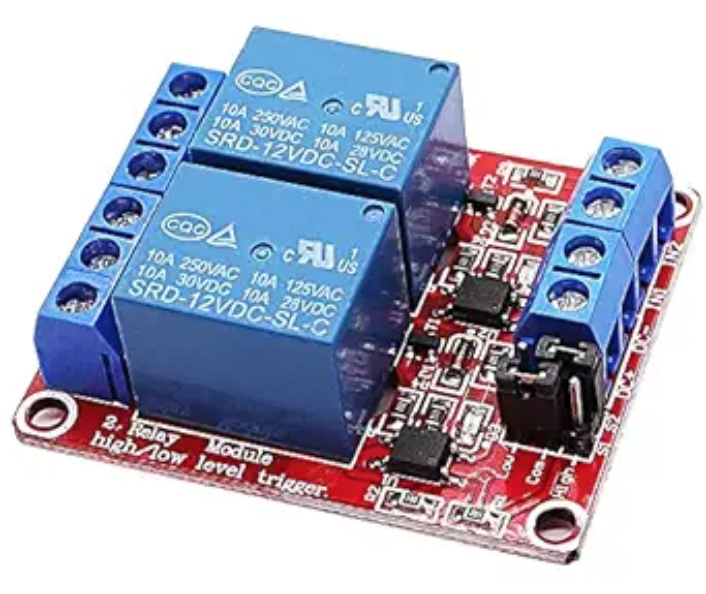} 
    \caption{12V DC relay switch}
    \label{fig:relay}
\end{figure}

\begin{figure}[H]
   \centering	\includegraphics[width=0.75\columnwidth]{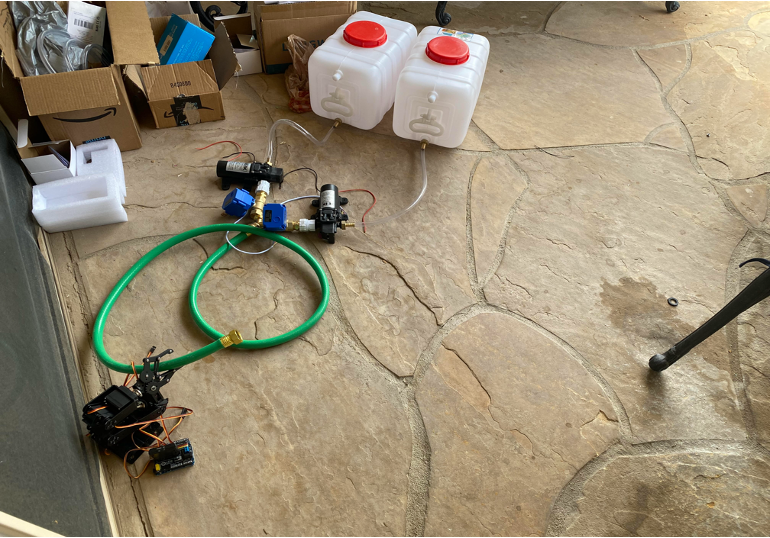} 
    \caption{Water Pumps and Tanks setup}
    \label{fig:setup}
\end{figure}

    The manipulator uses 6 DOF servo motors which consist of four MG996R servos in the revolute joints up to the wrist and two DS3218 servos used in the wrist and gripper.  Each of the two miniature diaphragm motor 12V 5.5A pumps (1.35 GPM.116 psi), shown in Figure \ref{fig:pump} is connected to a 3/4" 9-24V AC/DC electric motor valve ball shown in Figure \ref{fig:valves}. 
    Each of the motor valve balls opens and shuts the valve based on which of the water tank reservoir tanks is being used.  The water level is controlled using a  12V DC water level sensor controller module with automatic liquid control switch PCB Board that supports automatic water pumping/filling/draining.  When the water level of one tank gets too low, it switches tanks and the motor valve closes while the other opens. 
    The wiring of the motor pumps to the motor valve balls, shown in Figure \ref{fig:wiring} is controlled through a 2 Channel 12V DC relay switch shown in Figure \ref{fig:relay}.
   Thus, the robot actuators consist of the six manipulator servos, 4 DC planetary gear motors, two motor ball valve pumps, and two miniature diagram pumps which in turn are connected to 2 18L water tank reservoir tanks using brass adapters and connectors as illustrated in Figure \ref{fig:setup}.
    \ \ \ 

\section{Hardware Interfacing}

\ \ \ The overall architecture of the robot is shown in Appendix \ref{architecture}.
A PCA9685 AdaFruit controller board is connected to the 6 servos on the manipulator.  A 12V power supply battery goes through a step down converter which in turn sends +6V to the AdaFruit controller board.   This board is in turn connected to the Jetson Nano.  An Arduino Uno R3 microcontroller is also connected to the Jetson Nano.  It receives 12V from the power supply. Two L298 H-bridge circuit control boards are connected to the Arduino each of which controls 2 of the DC motors that power the chassis wheels.   
    
    The Jetson Nano is run with Robot Operating System 2 (ROS2) Humble. ROS 2 is a communication layer that allows different components of a robot system to exchange information in the form of messages. A component sends a message by publishing it to a particular topic, such as odometry. Other components receive the message by subscribing to that topic.  The manipulator and car chassis were designed and specified in a Unified Robot Description File (URDF), a file that contains a physical description and robot structure for the links and joints in an XML format.  The URDF tree for the manipulator is shown in Appendix \ref{urdf}.    

\section{Kinematics}

\subsection{Water Pressure and Fluid Dynamics}
\ \ \ A 6 DOF manipulator must hold and angle the hose at a height $h$ off the ground so that the water will reach the plant density from a distance $d$.  Figure \ref{fig:manipulator} shows an image of the 6 D0F manipulator:
\begin{figure}[H]
   \centering	\includegraphics[width=0.75\columnwidth]{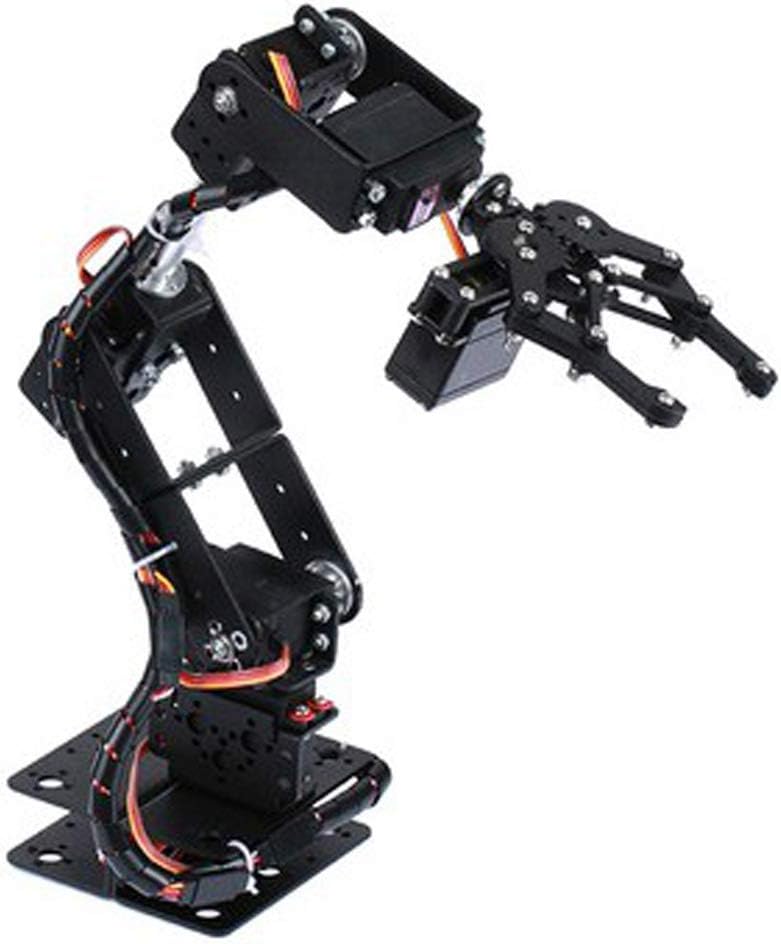} 
    \caption{6 DOF Manipulator}
    \label{fig:manipulator}
\end{figure} 
The height is equal to the height of the base of the chassis off the ground plus the vertical height of the manipulator.   The water passes through the hose at a speed/velocity $v$ based on the pounds per square inch.  The hose flow rate is expressed as gallons per minute (GPM).  

There are several factors that impact the flow rate including the diameter of the hose, the level of water pressure, and the length of the hose.   A typical garden hose rate is usually between 9 and 17 gallons per minute.  The average garden hose GPM is around 12 to 13, but can varying depending on the type of hose.   We use a garden hose with a diameter of 1/2".  Bernoulli's principle states that within a horizontal flow of fluid, points of higher fluid speed will have less pressure than powers of slower fluid speed. \cite{Bernard:2015} on Bernoulli's principle, when fluid flows through a pipe, the change of cross section area of the pipe causes change of velocity of fluid at that location, which causes change of pressure in the fluid.  

\begin{figure}[H]
   \centering	\includegraphics[width=0.75\columnwidth]{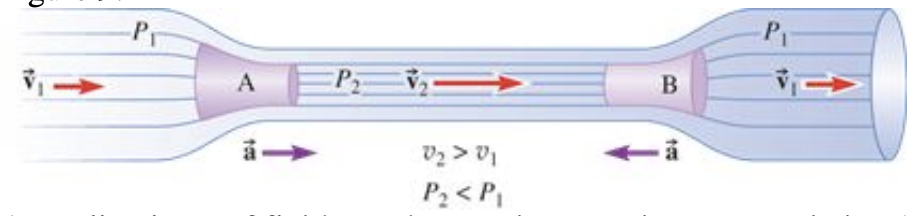} 
    \caption{Bernoulli's Principle. \cite{Giambattista:2020}} 
    \label{fig:bernoulli}
\end{figure}

Thus, the cross-sectional area of the nozzle determines the maximum velocity of the water exiting the nozzle.  Bernoulli's principle assumes the flow is steady, fluid is incompressible, there are no losses due to friction, no head addition or removal, and no pumps or motors that add or remove energy from the system.
The smaller the cross sectional area, the greater the velocity, as illustrated in Figure \ref{fig:bernoulli}.  
The pressure is the force divided by the cross sectional area, i.e., $P = F/A$.  The water flow rate $Q$ in gallons per minute (GPM) can be computed using the Hazen Williams equation:
\begin{equation}
    Q = \bigg (1946.6 \times D^{1.857} \times \frac{P}{L} \bigg )^{0.54}
\end{equation}
where $P$ is the pressure in PSI at the hose bib while hose the end is free flowing (dynamic pressure, not static pressure), $D$ is the hose internal diameter (in) and $L$ is the hose length.  The water flow rate is important is calculating how much water should be sprayed on the plant and therefore how long the robot should be spraying.  The water flow time is $T = \frac{Vol}{Q}$ where $Vol$ is the volume flow rate (volume velocity).  
 To calculate the water exit velocity from PSI and exit nozzle size, you can use the following formula: velocity = $\sqrt{(2 * PSI)}/Cd$ where $Cd$ is the discharge coefficient of the nozzle, which is a dimensionless number that depends on the shape of the nozzle and the ratio of the nozzle size to the pipe size. For a circular nozzle, the discharge coefficient is typically around 0.62 for a sharp-edged nozzle and around 0.97 for a fully-developed vena contracta nozzle.  For a 1/4" nozzle, the discharge coefficient is around 0.62. So, using the formula above, the water exit velocity for a 60 PSI pressure and a 1/4" nozzle would be v = $\sqrt{(2 * 60)}$ / 0.62 = 19.4 ft/s. This is approximate value and may vary depending on the actual conditions of the system, such as the presence of turbulence or the type of fluid being used.  The horizontal range of a projectile is
 $R = \frac{v^{2} \text{sin}(2\theta)}{g}$ where v is the initial velocity, $\theta$ is the angle at which the projectile is launched, and $g = 9.8 m/s^{2}$ is the acceleration due to gravity.  The maximum value of the sine function occurs at $90^{\degree}$, where it equals 1.  Thus, $\theta = 45^{\degree}$.   Thus, neglecting air resistance, water will travel farthest if aimed at $45^{\degree}$.  

 Exiting water that is sprayed from a nozzle $H$ ft above ground has both vertical and horizontal velocity, has a parabola trajectory:
 \begin{equation}
     y = H + x \bigg (\frac{v_{y,0}}{v_{x,0}} \bigg ) - x^{2} \bigg (\frac{g}{2v^{2}_{x,0}}\bigg )
 \end{equation}
where $v_{y,0}$ is the initial vertical velocity and $v_{x,0}$ is the initial horizontal velocity.  To calculate the position $(x_{1},y_{1},z_{1})$ that the water will reach in the $O_{1}$ frame we can use the general energy equation that is an extension of Bernoulli's equation to account for energy additions, removals and losses from pumps, motors and friction respectively.   

In terms of \textit{head} or the energy per unit weight of the fluid in $\frac{N \cdot m}{N}$ (SI) in the system $E$ that flows from point 1 to point 2 in the hose, 
the equation that conserves energy:
\begin{equation}
E = \frac{p_{1}}{\rho g} + z_{1} + \frac{v_{1}^{2}}{2g} + h_{A} - h_{R} - h_{L} = z_{2} + \frac{v_{2}^{2}}{2g} + \frac{p_{2}}{\rho g} 
\end{equation}
where $z$ is the elevation, $\rho$ is water density, $p$ is pressure, $g$ is the gravitational acceleration ($g = w/m$ : $w$ = fluid weight element and 
$m$ = fluid mass), $v$ is velocity, $h_{A}$ is the energy added to the fluid by pump, $h_{R}$ is the energy removed from fluid by a motor, and $h_{L}$ is the energy lost in the system due to friction. 

We assume the nozzle is located at $(x_{0},y_{0},z_{0})$ in the $O_{0}$ reference frame at the end of the gripper.  The manipulator is at an elevation $H$ off the ground.  However, we assume that the nozzle is aligned in the $z$-axis direction of the gripper, the same direction as the revolute joint attached to the gripper. 

\section{Forward Kinematics}
\begin{figure}[H]
   \centering	\includegraphics[width=0.75\columnwidth]{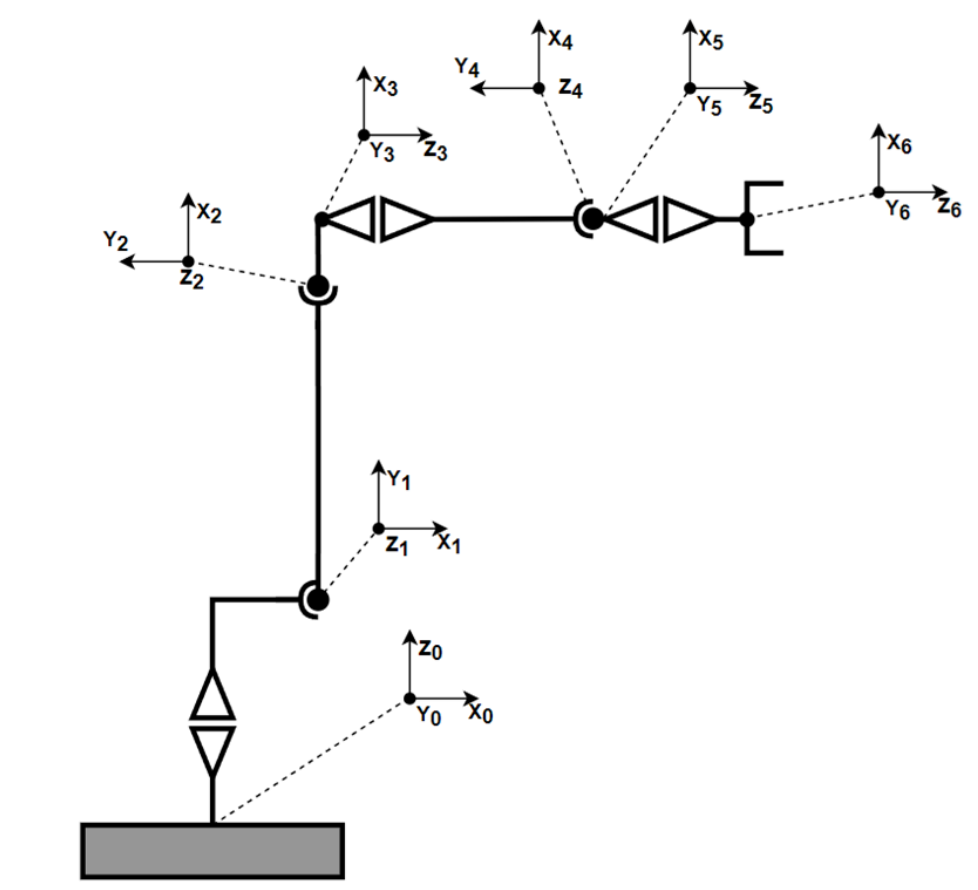} 
    \caption{Coordinate Frames}
    \label{fig:robot1}
\end{figure}

    The forward kinematics are similar to those of the ABB IRB arm 
    since both manipulators have 6 DOF and the same number of links and revolute joints in similar positions \cite{Cristoiu:2017, Seven:2019}.
    Figure \ref{fig:robot1} shows the coordinate frames at each of the joints.
    From the coordinates frames and link dimensions, a Uniform Robot Descriptor File (URDF) is created (see Appendix \ref{urdf}), the following DH Table for the manipulator was generated. 
\begin{center}
\begin{tabular}{|c c c c c c|} 
 \hline
 Order & $\theta$ & $\alpha$ & d  & a & Var \\ 
 \hline\hline
 1-2 & 0 & 0 & 0.0793 & 0 & $\theta_{1}$ \\
 2-3 & 0 & $\pi/2$ & 0.03 & 0 & $\theta_{2}$ \\ 
 3-4 & $-\pi/2$ & 0 & 0 & 0.127 & $\theta_{3}$ \\
 4-5 & 0 & 0 & 0 & 0.1842 & $\theta_{4}$ \\
 5-6 & $\pi/2$ & $\pi/2$ & 0 & 0 & $\theta_{5}$  \\
 6-7 & 0 & 0 & 0.1635 & 0 & $\theta_{6}$  \\
\hline
\end{tabular}
\end{center}
\ \ \ Appendix \ref{dh} includes a screen shot of the coordinate frames of the manipulator being shown in Matlab and the generation of the DH Table by hand and by computer.
\\
\\
    \ \ \ Let $a_{i}$, $\alpha_{i}$, $d_{i}$, and $\theta_{i}$ be the D-H parameters of link $i$, then a joint to joint homogeneous transformation matrix $\mathbf{T}^{i-1}_{i}$ which maps the coordinates from link $i-1$ to link $i$ can be defined based on four primary transformations:
\begin{equation}
    A_{i} = Rot(z,\theta_{i})Trans(z,d_{i})Trans(x,a_{i})Rot(x,\alpha_{i}) 
\end{equation}
where we can transfer from frame $i$ to frame $i+1$ frame using the following matrix transformations:
\begin{itemize}
    \item Rot(z,$\theta_{i}$) : rotation about z axis by angle $\theta_{i}$.
    \item Trans(z,$d_{i}$) : translation of z axis by distance $d_{i}$
    \item Trans(x,$a_{i}$) : translation of x axis by distance $a_{i}$ 
    \item Rot(x,$\alpha_{i}$) : rotation about x axis by angle $\alpha_{i}$ 
\end{itemize}
so that the final matrix is the result of their multiplication:
\begin{multline}
\mathbf{T}^{i-1}_{i} = 
         \begin{bmatrix}
                \text{cos}(\theta_{i}) & -\text{sin}(\theta_{i}) & 0 & 0 \\
                \text{sin}(\theta_{i}) & \text{cos}(\theta_{i}) & 0 & 0 \\
                0 & 0 & 1 & 0 \\
                0 & 0 & 0 & 1
            \end{bmatrix}
            \begin{bmatrix}
                1 & 0 & 0 & 0 \\
                0 & 1 & 0 & 0 \\
                0 & 0 & 0 & d_{i} \\
                0 & 0 & 0 & 1 
            \end{bmatrix} \cdot \\\\
            \begin{bmatrix}
               1 & 0 & 0 & a_{i} \\
               0 & 1 & 0 & 0 \\
               0 & 0 & 1 & 0 \\
               0 & 0 & 0 & 1 
            \end{bmatrix}
            \begin{bmatrix}
                1 & 0 & 0 & 0 \\
                0 & \text{cos}(\alpha_{i}) & -\text{sin}(\alpha_{i}) & 0 \\
                0 & \text{sin}(\alpha_{i}) & \text{cos}(\alpha_{i}) & 0 \\
                0 & 0 & 0 & 1 
            \end{bmatrix} \nonumber \\
    =
    \footnotesize
    \begin{bmatrix}
    \text{cos}(\theta_{i}) & -\text{cos}(\alpha_{i})\text{sin}(\theta_{i}) & \text{sin}(\alpha_{i})\text{sin}(\theta_{i}) & a_{i}\text{cos}(\theta_{i}) \\
    \text{sin}(\theta_{i}) & \text{cos}(\alpha_{i})\text{cos}(\theta_{i}) & -\text{sin}(\alpha_{i})\text{cos}(\theta_{i}) & a_{i}\text{sin}(\theta_{i}) \\
    0 & \text{sin}(\alpha_{i}) & \text{cos}(\alpha_{i}) & d_{i} \\
    0 & 0 & 0 & 1
    \end{bmatrix}    
\end{multline}
    \ \ \ The kinematic equation of a manipulator which completely describes the position and orientation of each link with respect to a base coordinate system \cite{Corke:2011, Craig:2004} can be expressed by the successive joint to joint mapping of adjacent links via
\begin{equation}
    \mathbf{T}^{0}_{i} = \prod_{j=1}^{i} \mathbf{T}^{j-1}_{j} \\\ \text{for} \ \ i = 1,2,\dots,n \ \ \ \nonumber \\
\end{equation}
\begin{align}
\mathbf{T}^{0}_{1} &=
\begin{bmatrix}
    \text{cos}(\theta_{1}) & -\text{sin}(\theta_{1}) & 0 & 0 \\
    \text{sin}(\theta_{1}) & \text{cos}(\theta_{1}) & 0 & 0  \\
    0 & 0 & 1 & d_{1} \\
    0 & 0 & 0 & 1
\end{bmatrix} \nonumber \\ 
\mathbf{T}^{1}_{2} &=
\begin{bmatrix}
    \text{cos}(\theta_{2}) & -\text{sin}(\theta_{2}) & 0 & 0\\
    \text{sin}(\theta_{2}) & \text{cos}(\theta_{2}) & 0 & 0 \\
    0 & 0 & 1 & d_{2} \\
    0 & 0 & 0 & 1
\end{bmatrix}  \nonumber \\
\mathbf{T}^{2}_{3} &=
\begin{bmatrix}
    \text{cos}(\theta_{3}) & 0 & -\text{sin}(\theta_{3}) & 0 \\
    \text{sin}(\theta_{3}) & 0 & \text{cos}(\theta_{3}) & 0  \\
    0 & 1 & 0 & d_{3} \\
    0 & 0 & 0 & 1
\end{bmatrix}  \nonumber \\
\mathbf{T}^{3}_{4} &=    
\footnotesize
\begin{bmatrix}
    \text{cos}(\theta_{4} - \pi/2) & -\text{sin}(\theta_{4} - \pi/2) & 0  & a_{4}\text{cos}(\theta_{4} - \pi/2) \\
    \text{sin}(\theta_{4} - \pi/2) & \text{cos}(\theta_{4} - \pi/2) & 0 & a_{4}\text{sin}(\theta_{4} - \pi/2) \\
    0 & 0 & 1 & d_{4} \\
    0 & 0 & 0 & 1 
\end{bmatrix} \nonumber \\
&=   
\begin{bmatrix}
    -\text{sin}(\theta_{4}) &  -\text{cos}(\theta_{4}) & 0 & -a_{4}\text{sin}(\theta_{4}) \\
    \text{cos}(\theta_{4}) & -\text{sin}(\theta_{4}) & 0 & a_{4}\text{cos}(\theta_{4})  \\
    0 & -1 & 0 & d_{3} \\
    0 & 0 & 0 & 1  \nonumber 
\end{bmatrix} \nonumber \\
\mathbf{T}^{4}_{5} &=    
\begin{bmatrix}
    \text{cos}(\theta_{5}) & 0 & \text{sin}(\theta_{5})  & a_{5}\text{cos}(\theta_{5}) \\
    \text{sin}(\theta_{5}) & 0 & -\text{cos}(\theta_{5}) & a_{5}\text{sin}(\theta_{5}) \\
    0 & 0 & 1 & d_{5} \\
    0 & 0 & 0 & 1 
\end{bmatrix} \nonumber \\
\mathbf{T}^{5}_{6} &=    
\begin{bmatrix}
    \text{cos}(\theta_{6} + \pi/2) & -\text{sin}(\theta_{6} + \pi/2) & 0  & 0 \\
    \text{sin}(\theta_{6} + \pi/2) & \text{cos}(\theta_{6} + \pi/2) & 0 & 0 \\
    0 & 0 & 1 & d_{6} \\
    0 & 0 & 0 & 1 
\end{bmatrix} \nonumber \\
&=
\begin{bmatrix}
    -\text{sin}(\theta_{6}) & -\text{cos}(\theta_{6}) & 0  & 0 \\
    \text{cos}(\theta_{6}) & -\text{sin}(\theta_{6}) & 0 & 0 \\
    0 & 0 & 1 & d_{6} \\
    0 & 0 & 0 & 1 
\end{bmatrix} \nonumber \\
\label{eq:trans}
\end{align}
    The total transformation matrix from the robot base to the end-effector is
\begin{equation}
    T_{f} = T^{0}_{1}T^{1}_{2}T^{2}_{3}T^{3}_{4}T^{4}_{5}T^{5}_{6} =
    \begin{bmatrix}
        n_{x} & o_{x} & a_{x} & p_{x} \\
        n_{y} & o_{y} & a_{y} & p_{y} \\
        n_{z} & o_{z} & a_{z} & p_{z} \\
        0 & 0 & 0 & 1
    \end{bmatrix}
\end{equation}
where $d_{2} = 0.0793, d_{3} = 0.03, d_{7} = 0.1635, a_{4} = 0.127, a_{5} = 0.1842$, and $d_{1} = d_{4} = d_{5} = d_{6} = 0$.
\begin{itemize}
\item $\textbf{n} = [n_{x} \ n_{y} \ n_{z}]$ is the unit vector in the direction of the X-axis at the robot end tip to the base coordinate system.
\item $\textbf{o} = [o_{x} \ o_{y} \ o_{z}]$ is the unit vector in the direction of the Y-axis at the robot end tip to the base coordinate system.  
\item $\textbf{a} = [a_{x} \ a_{y} \ a_{x}]$ is the unit vector in the direction of the Z-axis at the robot end tip to the base coordinate system. 
 \end{itemize}
Appendix \ref{elements} provides the transformation matrix elements computed by multiplying all the intermediate rotation and translation transformation matrices in equation \ref{eq:trans} each derived from $\mathbf{T}^{i-1}_{i}$.

\section{Inverse Kinematics}
\ \ \ We first provide the mathematical calculations to derive the inverse kinematics of the 6 DOF manipulator.  To express tip rotation in Euler angles ZY'Z".  Let the end-effector position of the 
\begin{itemize}
    \item $Z_{eff} = \text{arctan2} \big (\frac{a_{y}}{a_{x}} \big )$
    \item $Y'_{eff} = \text{arctan2} \bigg ( \frac{\sqrt{1-a^{2}_{z}}}{a_{z}} \bigg )$
    \item $Z"_{eff} = \text{arctan2} \bigg ( \frac{o_{z}}{-n_{z}} \bigg )$ 
\end{itemize}
    The transformation vector $\textbf{T}_{eff}$ includes the translation and ZY'Z" Euler angles can be represented as:
\begin{equation}
\begin{bmatrix}
    p_{x} \\
    p_{y} \\
    p_{z} \\
    Z_{eff} \\
    Y'_{eff} \\
    Z"_{eff}
\end{bmatrix} \nonumber
\end{equation}
We will use the geometric method to compute the theoretical inverse kinematic angles of the 6 DOF manipulator following \cite{Dikmentli:2022, Dhahari:2011, Seven:2019}.
\begin{figure}[H]
   \centering	\includegraphics[width=0.75\columnwidth]{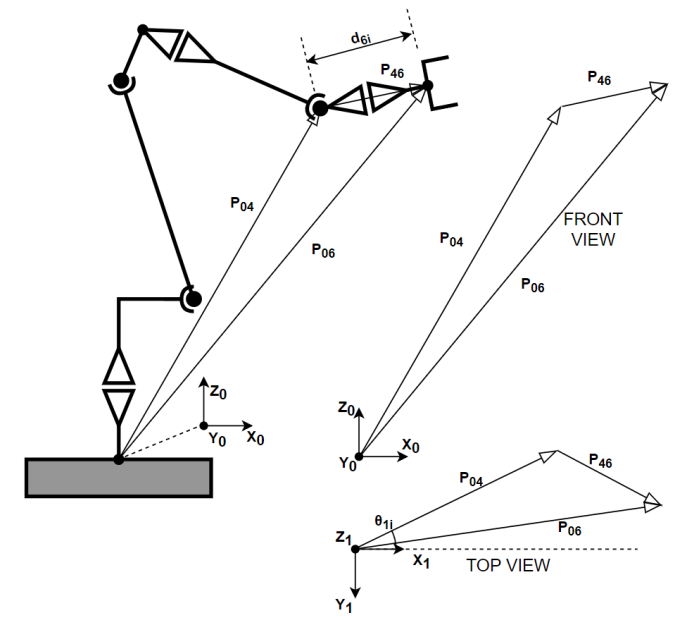} 
    \caption{Configuration to compute $\theta_{1}$. \cite{Dikmentli:2022}}
    \label{fig:robot3}
\end{figure} 
    Let $P_{ij}$ be the vector from the base coordinate frame $O_{i}$ to $O_{j}$, $0 \leq i < j \leq 6$.  Thus, 
    $P_{04} = P_{06} - P_{46}$
    where 
    \begin{equation}
    P_{06} = 
    \begin{bmatrix}
        p_{x6} \\ 
        p_{y6} \\
        p_{z6} \\
    \end{bmatrix}, \nonumber
    P_{46} = d_{6i}
    \begin{bmatrix} 
        a_{xi} \\
        a_{yi} \\
        a_{zi} \\
    \end{bmatrix}
    = \begin{bmatrix}
        d_{6i}a_{xi} \\
        d_{6i}a_{yi} \\
        d_{6i}a_{zi}
      \end{bmatrix}, \nonumber \\ 
      \end{equation}
      \begin{equation}
      P_{04} = \begin{bmatrix}
                    p_{04} \\
                    p_{04} \\
                    p_{04}
               \end{bmatrix}
               =
               \begin{bmatrix}
                p_{06} \\ 
                p_{06} \\
                p_{06} \\
               \end{bmatrix}     
                - \begin{bmatrix}
                d_{6i}a_{xi} \\
                d_{6i}a_{yi} \\
                d_{6i}a_{zi}
                \end{bmatrix}                 
                = 
                \begin{bmatrix}
                p_{06} - d_{6i}a_{xi} \\
                p_{06} - d_{6i}a_{yi} \\
                p_{06} - d_{6i}a_{zi}
                \end{bmatrix} \\
                \end{equation}.
                Therefore,
                $\theta_{1i} = \theta_{iv} = \text{arctan2}\bigg (\frac{p_{yi} - d_{6i}a_{yi}}{p_{xi} - d_{6i}a_{axi}} \bigg )$ \\
                $P_{01} =                 
                \begin{bmatrix}
                x_{01} \\
                y_{01} \\
                z_{01} 
                \end{bmatrix} 
                = \begin{bmatrix}
                   a_{1i}\text{cos}_{1i} \\ 
                   a_{1i}\text{sin}_{1i} \\
                   d_{1i}
                \end{bmatrix}$ and 
                $P_{04} =
                \begin{bmatrix}
                x_{04} \\
                y_{04} \\
                z_{04} 
                \end{bmatrix}                 
                = \begin{bmatrix}
                p_{xi} - d_{6i}a_{xi} \\
                p_{yi} - d_{6i}a_{yi} \\
                p_{zi} - d_{6i}a_{zi}
                \end{bmatrix}$.
\begin{figure}[H]
   \centering	\includegraphics[width=0.75\columnwidth]{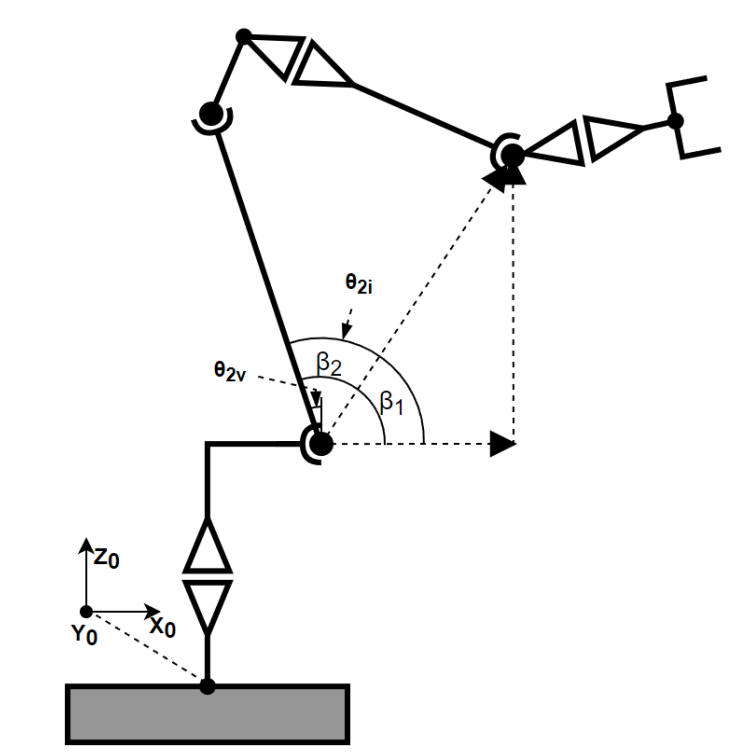} 
    \caption{Configuration to compute $\theta_{2}$. \cite{Dikmentli:2022}}
    \label{fig:robot5}
\end{figure} 
    From Figure \ref{fig:robot5}, $\beta_{1}$ and $\beta_{2}$ can be determined by:
\begin{align}
    \beta_{1} &= \text{tan}^{-1} \bigg ( \frac{z_{14}}{\sqrt{x^{2}_{14} + y^{2}_{14}}} \bigg ) \\
    \beta_{2} &= \text{tan}^{-1} \bigg ( \frac{a^{2}_{2i} - (P^{L}_{14})^{2} - l^{2}_{1}}{2a_{2}P^{L}_{14}} \bigg )
\end{align}
Therefore,
\begin{equation}
    \theta_{2i} = \theta_{2V} + \frac{\pi}{2} = \beta_{1} + \beta_{2}
\end{equation}
\begin{figure}[H]
   \centering	\includegraphics[width=0.75\columnwidth]{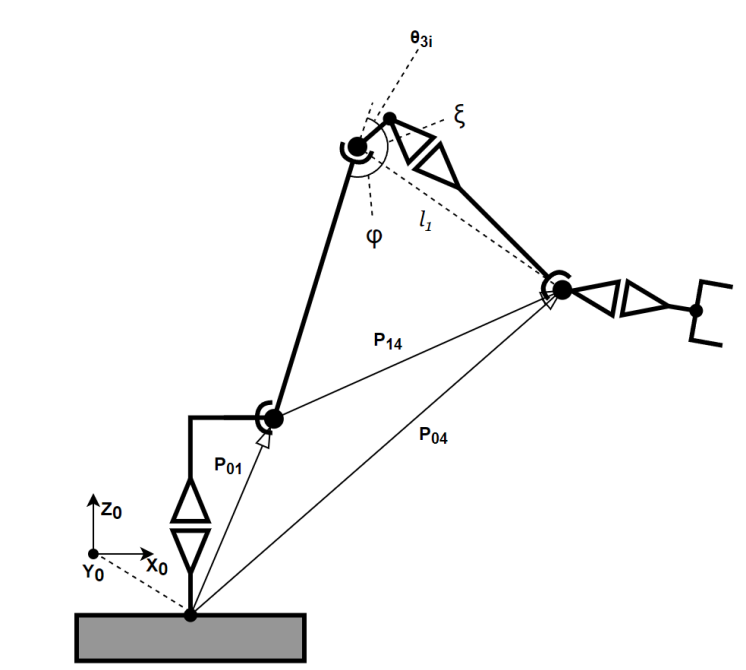} 
    \caption{Configuration to compute $\theta_{3}$. \cite{Dikmentli:2022}}
    \label{fig:robot4}
\end{figure} 
    From the law of cosines, 
\begin{equation}
    \text{cos}(\phi) = \frac{a^{2} + b^{2} - c^{2}}{2ab}
\end{equation}
Following from Figure \ref{fig:robot4}, 
$P_{14} = \begin{bmatrix}
            x_{14} \\
            y_{14} \\
            z_{14}
          \end{bmatrix}
          = P_{04} - P_{01} = 
          \begin{bmatrix}
            p_{xi} - d_{6i}a_{xi} - a_{1i}\text{cos}_{1i}(\alpha_{i}) \\
            p_{yi} - d_{6i}a_{yi} - a_{1i}\text{sin}_{1i}(\alpha_{i}) \\
            p_{zi} - d_{6i}a_{zi} - d_{1i}
          \end{bmatrix}$ 
    The length of $P^{L}_{14} = \lvert \rvert P^{L}_{14} \rvert \lvert = \sqrt{\lvert P_{14}^{T} \cdot P_{14} \rvert}, \ l_{1} = \sqrt{a^{2}_{3i} + d^{2}_{4i}}$
    where the superscript $L$ denotes the length of the vector. Applying the law of cosines and trigonometric formulas:
    \begin{equation}
        \phi = \text{cos}^{-1} \bigg (\frac{l^{2}_{1} + a^{2}_{2i} - P^{L}_{14}}{2l_{1}a_{2}} \bigg ), \ \ \ \zeta = \text{tan}^{-1} \bigg (\frac{d_{4i}}{a_{3i}} \bigg ) \nonumber
    \end{equation}
    Therefore, 
    \begin{equation}
        \theta_{3i} = \phi - \zeta - \pi
    \end{equation}
\begin{figure}[H]
   \centering	\includegraphics[width=0.75\columnwidth]{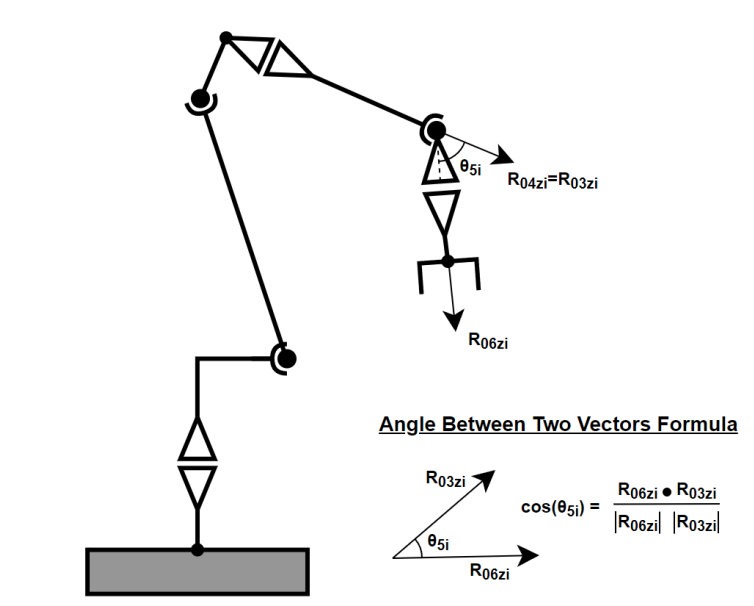} 
    \caption{Configuration to compute $\theta_{5}$.  \cite{Dikmentli:2022}}
    \label{fig:robot6}
\end{figure}
    Assume $\theta_{4i} = 0$.  Then the angle between the rotational vectors ($R_{z}$) will yield $\theta_{5i}$.
    Define the following transformation matrices:
    \begin{align}
    T_{1i} &= \begin{bmatrix}
        \text{cos}(\alpha_{1i}) & 0 & \text{sin}(\alpha_{1i}) & a_{1i}\text{cos}(\alpha_{1i}) \\
        \text{sin}(\alpha_{1i}) & 0 & -\text{cos}(\alpha_{1i}) & a_{1i}\text{sin}(\alpha_{1i}) \\
        0 & 1 & 0 & d_{1i} \\
        0 & 0 & 0 & 1
    \end{bmatrix} \nonumber \\
     T_{2i} &= 
     \begin{bmatrix}
        \text{cos}(\beta_{2i}) & -\text{sin}(\beta_{2i}) & 0 & a_{2i}\text{cos}(\beta_{2i}) \\
        \text{sin}(\beta_{2i}) & \text{cos}(\beta_{2i}) & 0 & a_{2i}\text{sin}(\beta_{2i}) \\
        0 & 0 & 1 & 0 \\
        0 & 0 & 0 & 1
    \end{bmatrix} \nonumber \\
     T_{3i} &= 
     \begin{bmatrix}
        \text{cos}(\gamma_{3i}) & 0 & \text{sin}(\gamma_{3i}) & a_{3i}\text{cos}(\gamma_{3i}) \nonumber \\
        \text{sin}(\gamma_{3i}) & 0 & -\text{cos}(\gamma_{3i}) & a_{3i}\text{sin}(\gamma_{3i}) \nonumber \\
        0 & 1 & 0 & 0 \\
        0 & 0 & 0 & 1
    \end{bmatrix} \nonumber
    \end{align}
    From the trigonometric identities for cosines and sines, we know:
    \begin{align} 
        \text{cos}(\alpha_{1i} + \beta_{2i}) &= \text{cos}(\alpha_{1i})\text{cos}(\beta_{2i}) - \text{sin}(\alpha_{1i})\text{sin}(\beta_{2i}) \nonumber \\
        \text{sin}(\alpha_{1i} + \beta_{2i}) &= \text{cos}(\alpha_{1i})\text{sin}(\beta_{2i}) + \text{sin}(\alpha_{1i})\text{cos}(\beta_{2i}) \nonumber
    \end{align} 
    We have assumed the 4th joint is absent for the purpose of finding the 5th joint angle.  The first and third joint transformation matrices will be considered to be the first to the fourth joint transformation matrices.  Since we know the first three joint angles, $T_{13}$ and subsequently $T_{14}$ can be derived as follows:
    $T_{14i} = T_{13i} = T_{1i} \cdot T_{2i} \cdot T_{3i} = $
    \begin{equation}
    \begin{bmatrix}
    I_{x} & J_{x} & K_{x} & L_{x} \\
    I_{y} & J_{y} & K_{y} & L_{y} \\
    I_{z} & J_{z} & K_{z} & L_{z} \\
    0 & 0 & 0 & 1
    \end{bmatrix} 
    \end{equation}
    where 
    \begin{align}
    I_{x} &= \text{C}(\alpha_{1i})\text{C}(\beta_{2i} + \gamma_{3i}) \nonumber \\ 
    J_{x} &= S(\alpha_{1i}) \nonumber \\
    K_{x} &= C(\alpha_{1i})S(\beta_{2i} + \gamma_{3i}) \nonumber \\
    L_{x} &= \text{C}(\alpha_{1i})(a_{1i} + a_{2i}\text{C}(\beta_{2i}) + a_{3i}\text{C}(\beta_{2i} + \gamma_{3i})) \nonumber \\ 
    I_{y}  &= \text{S}(\alpha_{1i})\text{C}(\beta_{2i} + \gamma_{3i}) \nonumber \\
    J_{y} &= -C(\alpha_{1i}) \nonumber \\
    K_{y} &= \text{S}(\alpha_{1i})\text{S}(\beta_{2i} + \gamma_{3i}) \nonumber \\  
    L_{y} &= \text{S}(\alpha_{1i})(a_{1i} + a_{2i}\text{C}(\beta_{2i}) + a_{3i}\text{C}(\beta_{2i} + \gamma_{3i})) \nonumber \\
    I_{z} &= \text{S}(\beta_{2i} + \gamma_{3i})  \nonumber \\
    J_{z} &=  0  \nonumber \\ 
    K_{z} &=  -\text{C}(\beta_{2i} + \gamma_{3i}) \nonumber \\ 
    L_{z} &=  a_{2i}\text{S}(\beta_{2i}) + d_{1i} + a_{3i}\text{S}(\beta_{2i} + \gamma_{3i}) \nonumber 
    \end{align}
Given input values, the rotation vector about z ($R_{6zi}$) is known.  From the above assumption that $R_{4zi}$ is equal to $R_{3zi}$, $R_{4zi}$ is known.  Then the angle between two vectors formula can be applied to find the unknown angle $\theta_{5i}$.
\begin{equation}
    R_{6zi} = 
    \begin{bmatrix}
        a_{xi} \\
        a_{yi} \\
        a_{zi}
    \end{bmatrix},
    \ \ \ R_{3zi} = R_{4zi} = \begin{bmatrix}
                                \text{cos}(\alpha_{1i})\text{sin}(\beta_{2i} + \gamma_{3i}) \\
                                \text{sin}(\alpha_{1i})\text{sin}(\beta_{2i} + \gamma_{3i}) \\
                                -\text{cos}(\beta_{2i} + \gamma_{3i})
                              \end{bmatrix} \nonumber
\end{equation}
    Since both $R_{6zi}$ and $R_{4zi}$ are unit vectors, matrix multiplication will yield the cosine of the angle between two vectors:
\vspace{-2mm}
\begin{equation}
    \theta_{5i} = \theta_{5V} = \text{cos}^{-1}(\textbf{R}_{6zi} \cdot \textbf{R}_{3zi})
\end{equation}
    The fourth and sixth joints are the only joint angles that are still unknown.  Since both $T_{3i}$ and $T_{6i}$ are known, one can derive $T_{46i}$ as: 
    \begin{equation}
        T_{46i} = T^{-1}_{13i} \cdot T_{16i} =
        \begin{bmatrix}
            I_{x} & J_{x} & K_{x} & L_{x} \\
            I_{y} & J_{y} & K_{y} & L_{y} \\
            I_{z} & J_{y} & K_{z} & L_{z} \\
            0 & 0 & 0 & 1  \nonumber
        \end{bmatrix}
    \end{equation}
    On the other hand, the matrix $T_{46i}$ can also be derived symbolically the multiplying the following matrices: \\
    $T_{46i} = T_{4i} \cdot T_{5i} \cdot T_{6i} = $
\begin{equation}
\begin{bmatrix}
    E_{x} & F_{x} & G_{x} & H_{x} \\
    E_{y} & F_{y} & G_{y} & H_{y} \\
    E_{z} & F_{z} & G_{z} & H_{z} \\
    0 & 0 & 0 & 1 \nonumber
\end{bmatrix}
\end{equation}
where
\begin{align}
    E_{x} &= C_{4i}C_{5i}C_{6i} - S_{4i}S_{6i} \nonumber \\ 
    F_{x} &= -(C_{4i}C_{5i}S_{6i} + S_{4i}C_{6i}) \nonumber \\ 
    G_{x} &= C_{4i}S_{5i} \nonumber \\ 
    H_{x} &= C_{4i}S_{5i}d_{6i} \nonumber \\
    E_{y} &= S_{4i}C_{5i}C_{6i} + C_{4i}S_{6i} \nonumber \\ 
    F_{y} &= -S_{4i}C_{5i}S_{6i} + C_{4i}C_{6i} \nonumber \\ 
    G_{y} &= S_{4i}S_{5i} \nonumber \\ 
    H_{y} &= S_{4i}S_{5i}d_{6i} \nonumber \\
    E_{z} &= -S_{5i}C_{6i} \nonumber \\ 
    F_{z} &= S_{5i}S_{6i} \nonumber \\ 
    G_{z} &= C_{5i}  \nonumber \\
    H_{z} &= C_{5i}d_{6i} + d_{4i} \nonumber
\end{align} 
    From this matrix, the inverse tangent (arctan2) of matrix element (2,3) ($G_{y}$) over element (1,3) ($G_{x}$) gives $\theta_{4i}$:
\begin{equation}
\theta_{4i} = \theta_{4v} = \text{arctan2} \bigg (\frac{G_{y}}{G_{x}} \bigg )
\end{equation}
    The inverse tangent of element (2,3) ($F_{z}$) over element (3,1) ($E_{z}$) gives $\theta_{6i}$.
\begin{equation}
\theta_{6i} = \theta_{6v} = \text{arctan2} \bigg (\frac{-F_{z}}{E_{z}} \bigg )
\end{equation}
\section{Simulation of Inverse Kinematics}
\begin{figure}[H]
   \centering	\includegraphics[width=0.75\columnwidth]{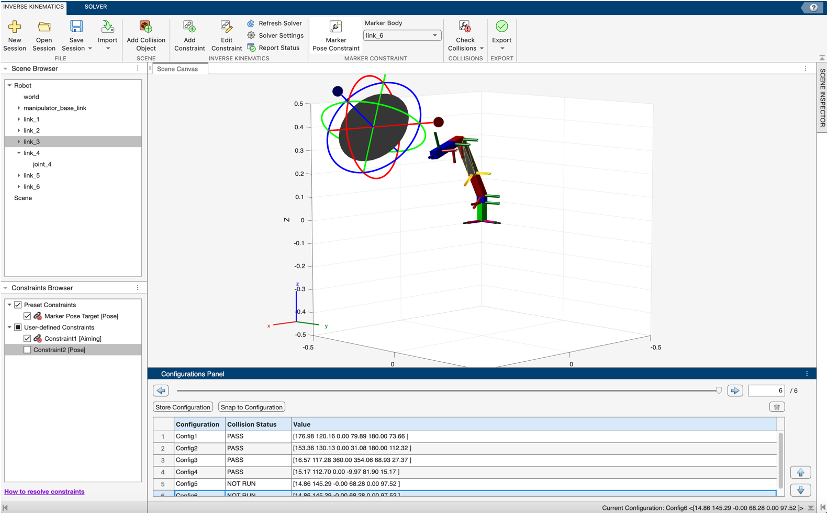} 
    \caption{Inverse Kinematics for Pose 1} 
    \label{fig:matlab1}
\end{figure}

\begin{figure}[H]
   \centering	\includegraphics[width=0.75\columnwidth]{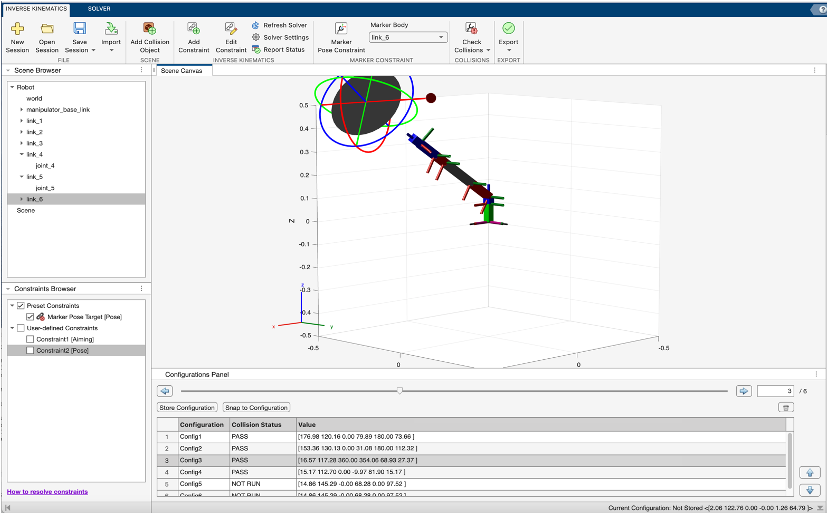} 
    \caption{Inverse Kinematics for Pose 2} 
    \label{fig:matlab2}
\end{figure}

\begin{figure}[H]
   \centering	\includegraphics[width=0.75\columnwidth]{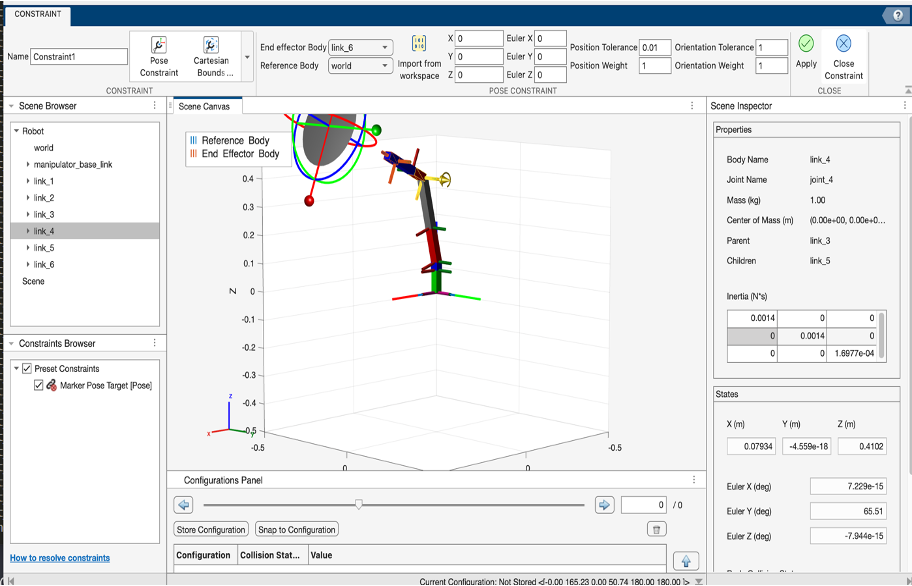} 
    \caption{Inverse Kinematics for Pose 3} 
    \label{fig:matlab4}
\end{figure}

\begin{figure}[H]
   \centering	
\includegraphics[width=0.75\columnwidth]{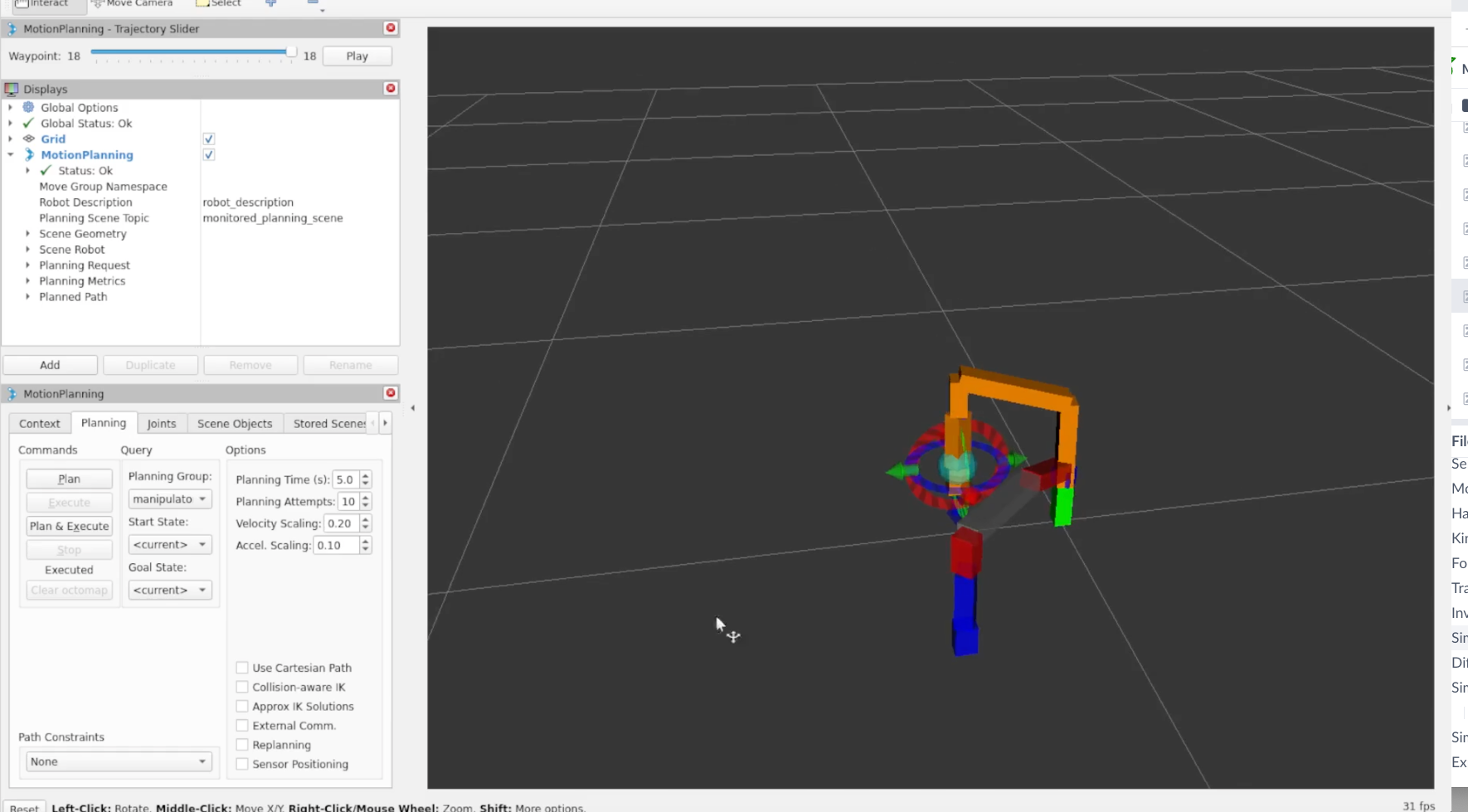} 
    \caption{MoveIt Pose 1} 
    \label{fig:moveit1}
\end{figure}

\begin{figure}[H]
   \centering	\includegraphics[width=0.75\columnwidth]{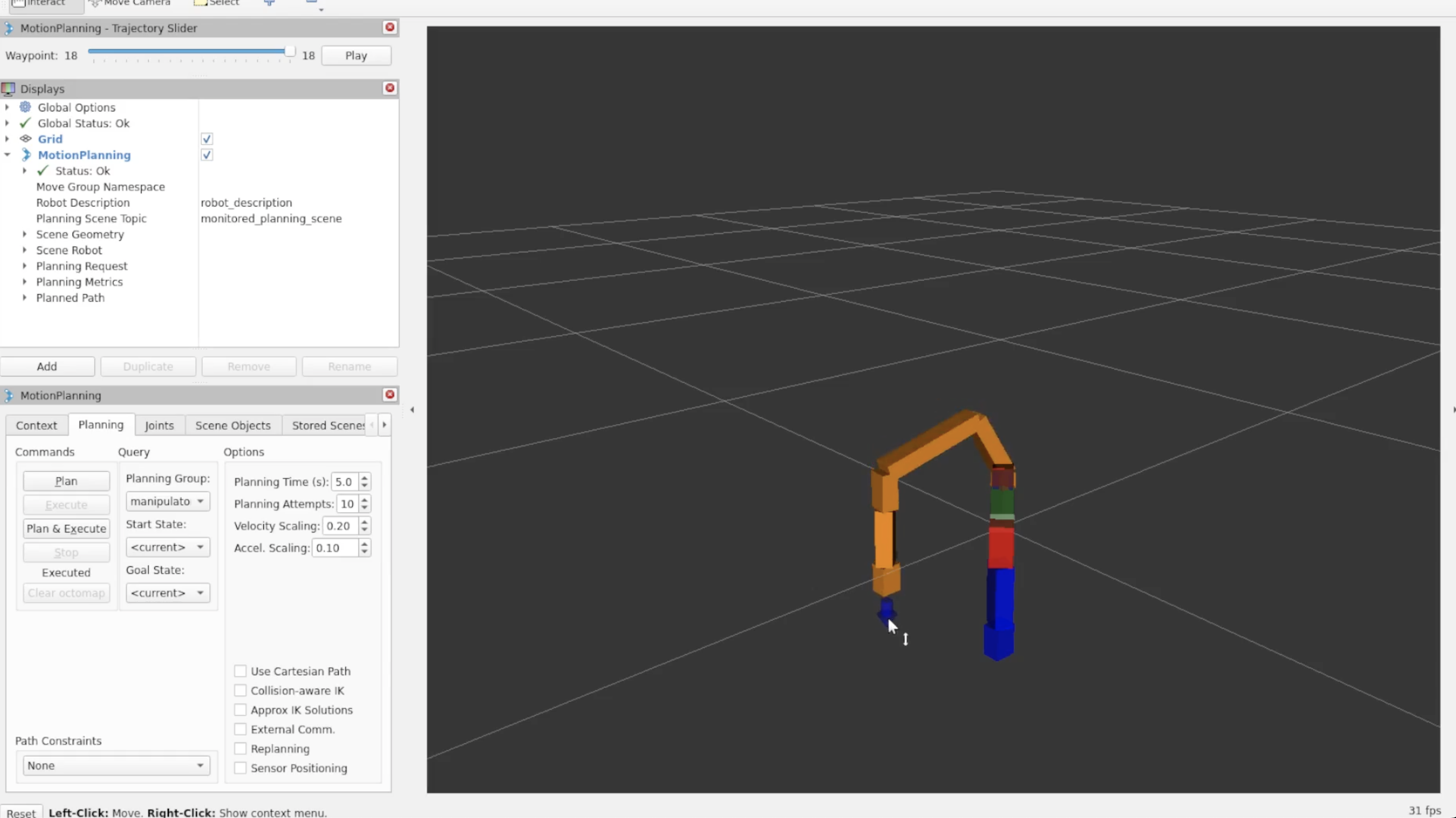} 
    \caption{MoveIt Pose 2} 
    \label{fig:moveit2}
\end{figure}

\begin{figure}[H]
   \centering	\includegraphics[width=0.75\columnwidth]{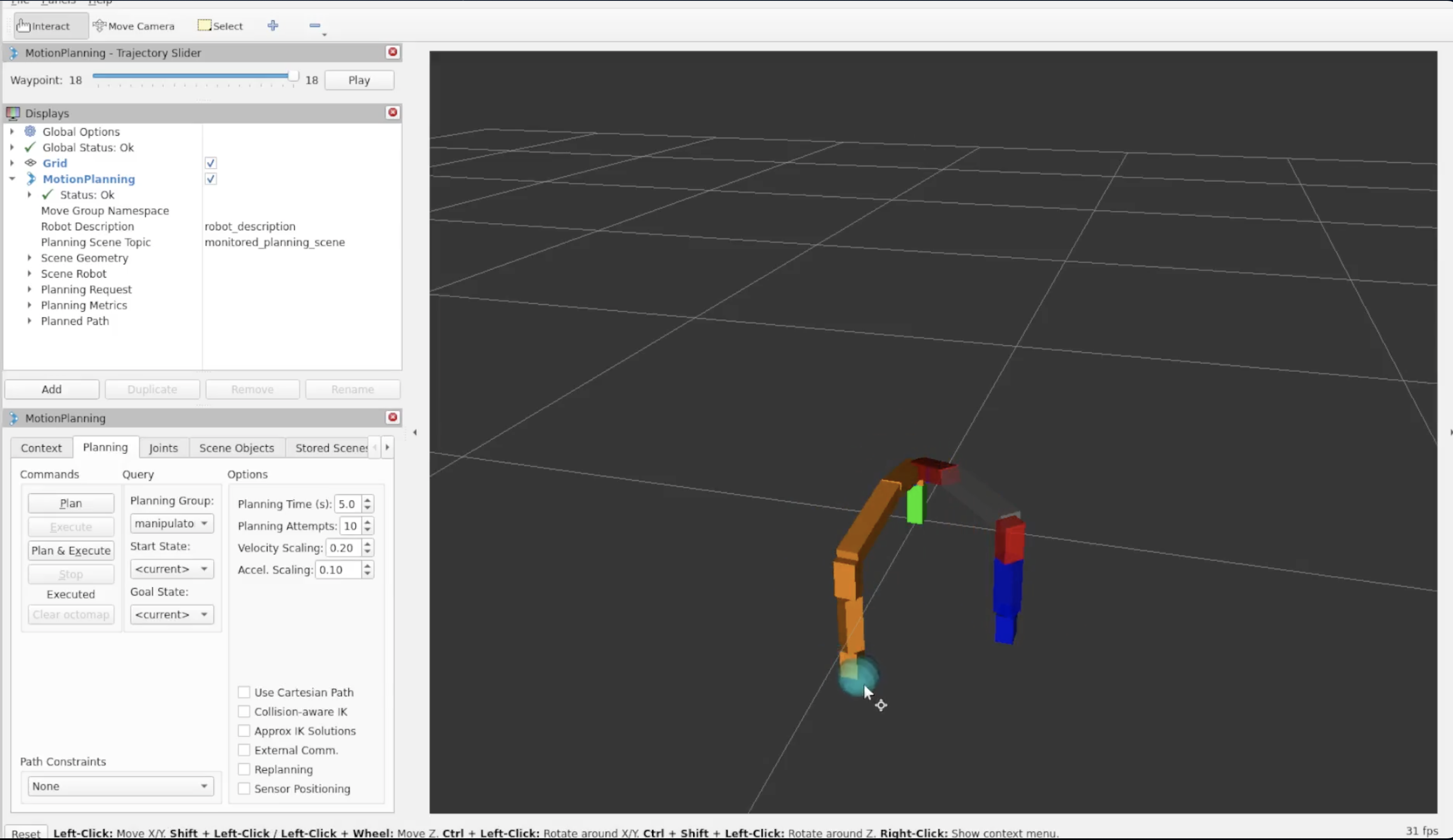} 
    \caption{Moveit Pose 3} 
    \label{fig:moveit3}
\end{figure}

\ \ \ We also use the Matlab Inverse Kinematics Toolkit.  The robot's URDF is first imported and through Matlab's inverse solver (such as Levenberg-Marquardt and Gauss-Newton), the six joint angles are determined by the selected pose of the manipulator.  The inverse kinematics of various poses are shown in Figures \ref{fig:matlab1}, \ref{fig:matlab2}, and \ref{fig:matlab4}.  The six angles generated from the inverse kinematics solver are shown at the bottom of each figure.
Using MoveIt, a motion planning framework for ROS, we simulate the inverse kinematic motions as shown in Figures \ref{fig:moveit1}, \ref{fig:moveit2}, and \ref{fig:moveit3} using different poses.\footnote{\url{moveit.ai}}  
\section{Differential Driving Kinematics}
\begin{figure}[H]
   \centering	\includegraphics[width=0.6\columnwidth]{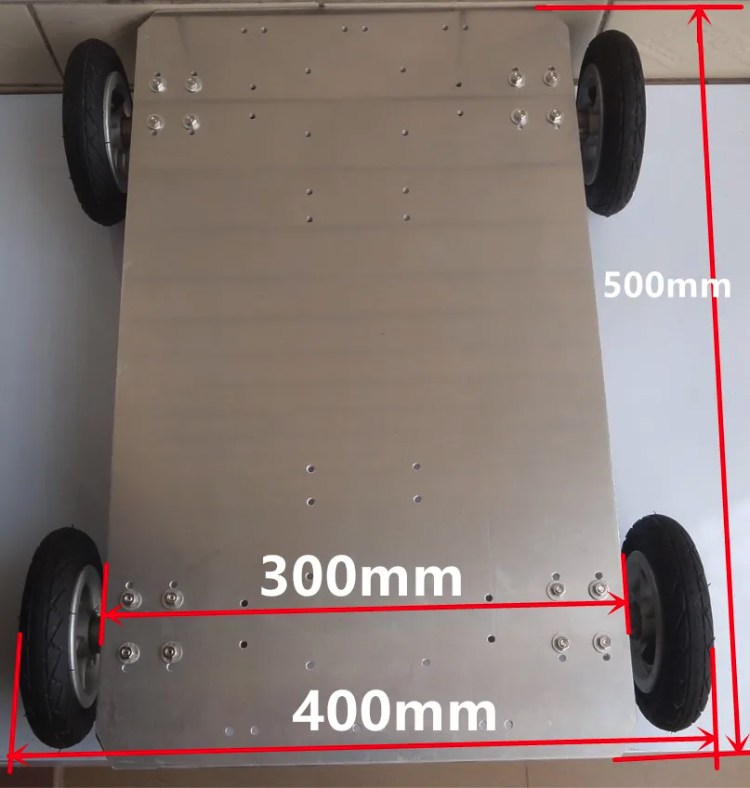} 
    \caption{Chassis (top view)}
    \label{fig:car2}
\end{figure}

\begin{figure}[H]
   \centering	\includegraphics[width=0.6\columnwidth]{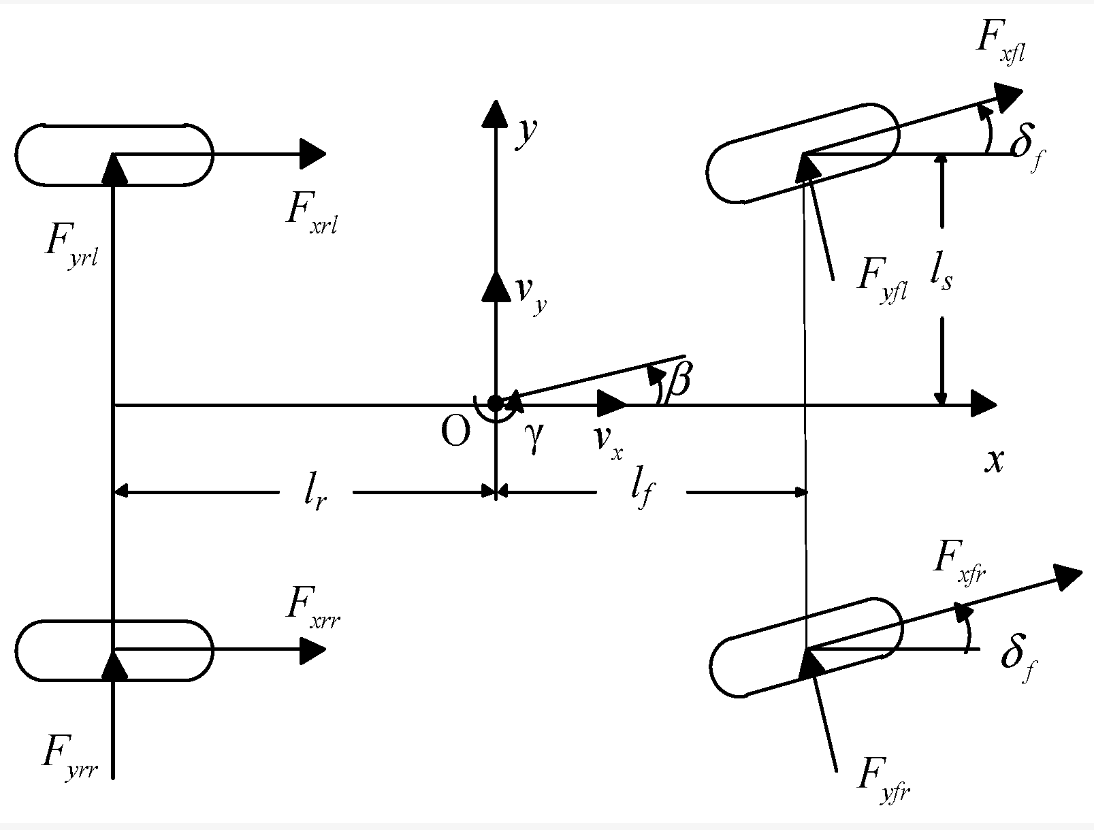} 
    \caption{Dynamic model of DSV}
    \label{fig:diffdrive}
    \end{figure}

  \ \ \ Figure \ref{fig:car2} shows a top view of the steel 4 wheel drive chassis. The chassis has a length of 500 mm and a width of 300 mm.  Unfortunately, due to prolonged delayed in receiving the chassis from the vendor in China, only simulation using the chassis were able to be performed rather than real experiments.
    Figure \ref{fig:diffdrive} shows the forces acting on the body of a differential steering vehicle (DSV) where $F_{xij}(i=f,r,j=l,r)$ is the tire longitudinal force on the left (l)/right (r) front (f)/rear wheel (r), $F_{yij}(i=f,r,j=l,r)$ is the tire lateral force of the left/right front/rear wheel.  The electronic control unit (ECU) provides instructions to the steering mechanism and achieves the steering according to the collected signals and internal control procedures.  For instance, when LIDAR detects an object that the robot car may hit, the ECU will receive the coordinates of that object from the LIDAR signal so that it can steer the car to a path away to avoid a collision.
    Following Tian et. al. \cite{Tian:2018}, the lateral and yaw motion of the DSV are modeled as: 
\begin{multline}
    m u_{x}(\dot{\beta} + \gamma) = (F_{yfl} + F_{yfr})\text{cos}(\delta_{f}) + \\ (F_{xfl} + F_{xfr})\text{sin}(\delta_{f}) + F_{yrl} + F_{yrr} 
    I_{z}\dot{\gamma} \\ = (l_{f}\text{sin}(\delta_{f}) + l_{s}\text{cos}(\delta_{f}))\frac{\Delta M}{R} \\
    + (l_{f}\text{cos}(\delta_{f}) - l_{s}\text{sin})(\delta_{f})(F_{yfl} + F_{yfr}) - l_{r}(F_{yrl} + F_{yrr})' \nonumber
\end{multline} 
where
\begin{align}
    F_{yfl} &= F_{yfr} = k_{f}(\beta + l_{f}\gamma/u_{x} - \delta_{f}) \nonumber \\
    F_{yrl} &= F_{yrr} = k_{r}(\beta - l_{r}\gamma/u_{x}) \nonumber 
\end{align}
    \begin{itemize}
    \item $m$ is the total vehicle mass
    \item $u_{x}$ is the longitudinal velocity at the center of gravity (CG) point
    \item $\beta$ is the slipside angle
    \item $\gamma$ is the yaw rate, 
    \item $I_{z}$ is the yaw moment of inertia 
    \item $I_{s}$ is the half of wheel track
    \item R is the radius of the front wheel 
    \item $l_{f}$ and $l_{r}$ are the distances from thh CG to the front and right axles 
    \item $\Delta M$ is the differential driving torque between the two sides of the front wheels
    \end{itemize}
    The mass $m$ includes not only the chassis weight, but also the weight of the two 18L water reservoir tanks that weigh 39.68 lbs each when full, or 79.37 lbs total, as well as the weight of the water motor pumps, motor valve balls, hose, manipulator, and other components attached and being carried top of the chassis.

    The dynamic equations of the steering systems are:
\begin{align}
    J_{e}\ddot{\delta}_{f} + b_{e}\dot{\delta}_{f} &= \tau_{a} + \frac{\Delta M}{R}r_{\sigma} - \tau_{f} \\ 
    \tau_{a} &= k_{f}\alpha_{f}l^{2}/3 \\
    \alpha_{f} &= \beta + l_{f}\gamma/u_{x} - \delta_{f}
\end{align}
 where 
 \begin{itemize}
    \item $J_{e}$ is the effective moment of inertia
    \item $b_{e}$ is the damping of the steer-by-wire (SBW) system
    \item $\delta_{f}$ is the front wheel steering angle
    \item $\tau_{a}$ is the tire self-aligning moment
    \item $r_{\sigma}$ is the scrub radius
    \item $\tau_{f}$ is the friction torque
    \item $\alpha_{f}$ is the tire slip angle of the front wheel
    \item $l$ is half of the tire contact length
 \end{itemize}
 \ \ \ Let $\textbf{X}(t) = [\beta \ \gamma \ \delta_{f}]^{T}$, $U(t) = \Delta M$.  Then the state equation is 
 \begin{equation}
        \dot{X} = A_{s}X + B_{s}U \nonumber
 \end{equation}
 where 
 \begin{equation}
    A_{s} = \begin{bmatrix}
            \frac{2k_{f} + 2k_{r}}{mu_{x}} & \frac{2k_{f}l_{f} - 2k_{r}l_{r}}{mu^{2}_{x}} - 1 & \frac{2k_{f}}{mu_{x}} \\ 
            \frac{2k_{f}l_{f} - 2k_{r}l_{r}}{I_{Z}}& \frac{2k_{f}l^{2}_{f} - 2k_{r}l^{2}_{r}}{I_{Z}u_{x}} & -\frac{2k_{f}l_{f}}{I_{Z}} \\ 
            \frac{k_{f}l^{2}}{2b_{e}} & \frac{k_{f}l^{2}l_{f}}{3b_{e}u_{x}} & -\frac{k_{f}l^{2}}{3b_{e}}
            \end{bmatrix}  \nonumber
 \end{equation}
 and 
 \begin{equation}
    B_{s} = \begin{bmatrix}
                0 \\
                \frac{l_{s}}{I_{Z}R} \\
                \frac{r_{\sigma}}{Rb_{e}}
            \end{bmatrix} \nonumber
\end{equation}
Thus,
\vspace{-2mm}
\begin{multline}
\begin{bmatrix}
    \dot{\beta} \\
    \dot{\gamma} \\
    \dot{\delta_{f}}
\end{bmatrix} = \\
    \begin{bmatrix}
            \frac{2k_{f} + 2k_{r}}{mu_{x}} & \frac{2k_{f}l_{f} - 2k_{r}l_{r}}{mu^{2}_{x}} - 1 & \frac{2k_{f}}{mu_{x}} \\ 
            \frac{2k_{f}l_{f} - 2k_{r}l_{r}}{I_{Z}}& \frac{2k_{f}l^{2}_{f} - 2k_{r}l^{2}_{r}}{I_{Z}u_{x}} & -\frac{2k_{f}l_{f}}{I_{Z}} \\ 
            \frac{k_{f}l^{2}}{2b_{e}} & \frac{k_{f}l^{2}l_{f}}{3b_{e}u_{x}} & -\frac{k_{f}l^{2}}{3b_{e}}
    \end{bmatrix}
  \begin{bmatrix}
    \beta \\
    \gamma \\
    \delta_{f}   
    \end{bmatrix} \\
+ \begin{bmatrix}
0 \\
\frac{l_{s}}{I_{Z}R} \\
\frac{r_{\sigma}}{Rb_{e}}
\end{bmatrix} 
\Delta M  \nonumber
\end{multline}

\section{Simulation}
    


\begin{figure}[H]
   \centering	
\includegraphics[width=0.75\columnwidth]{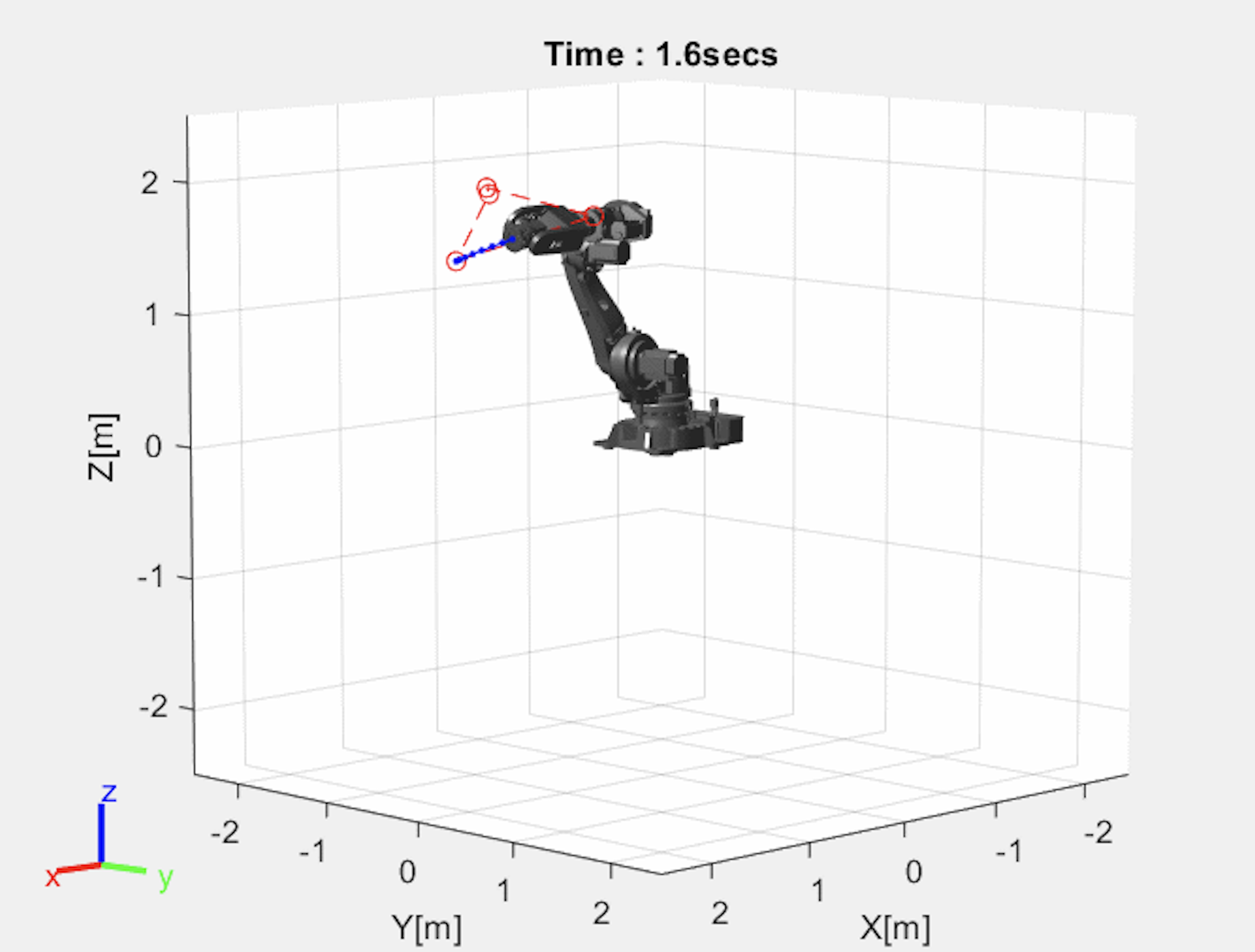} 
    \caption{Manipulator Workspace Simulation 1} 
    \label{fig:sim0}
\end{figure}

\begin{figure}[H]
   \centering	
\includegraphics[width=0.75\columnwidth]{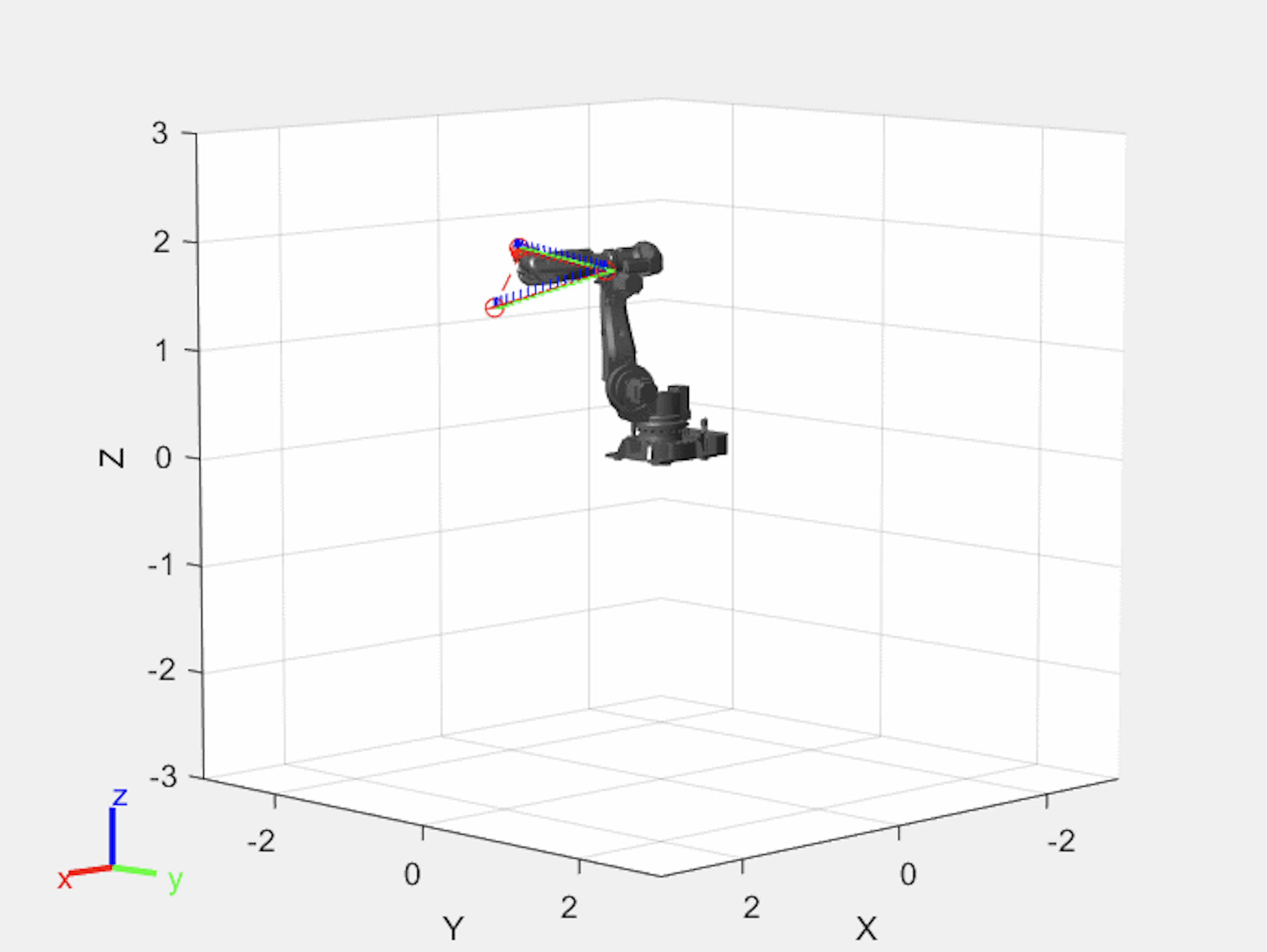} 
    \caption{Manipulator Workspace Simulation 2} 
    \label{fig:sim1}
\end{figure}

\begin{figure}[H]
   \centering	
\includegraphics[width=0.75\columnwidth]{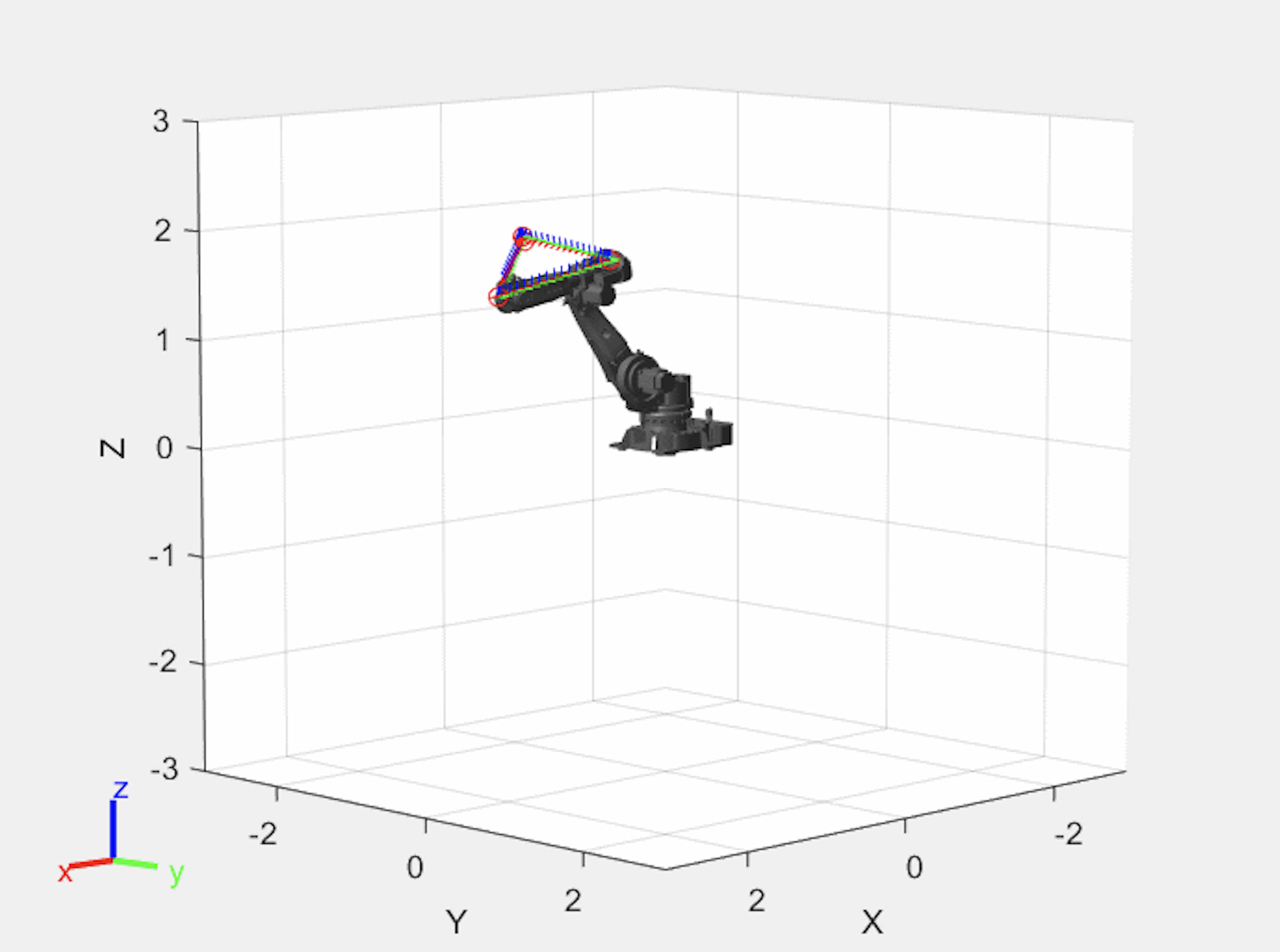} 
    \caption{Manipulator Workspace Simulation 3} 
    \label{fig:sim2}
\end{figure}

\begin{figure}[H]
   \centering	\includegraphics[width=0.75\columnwidth]{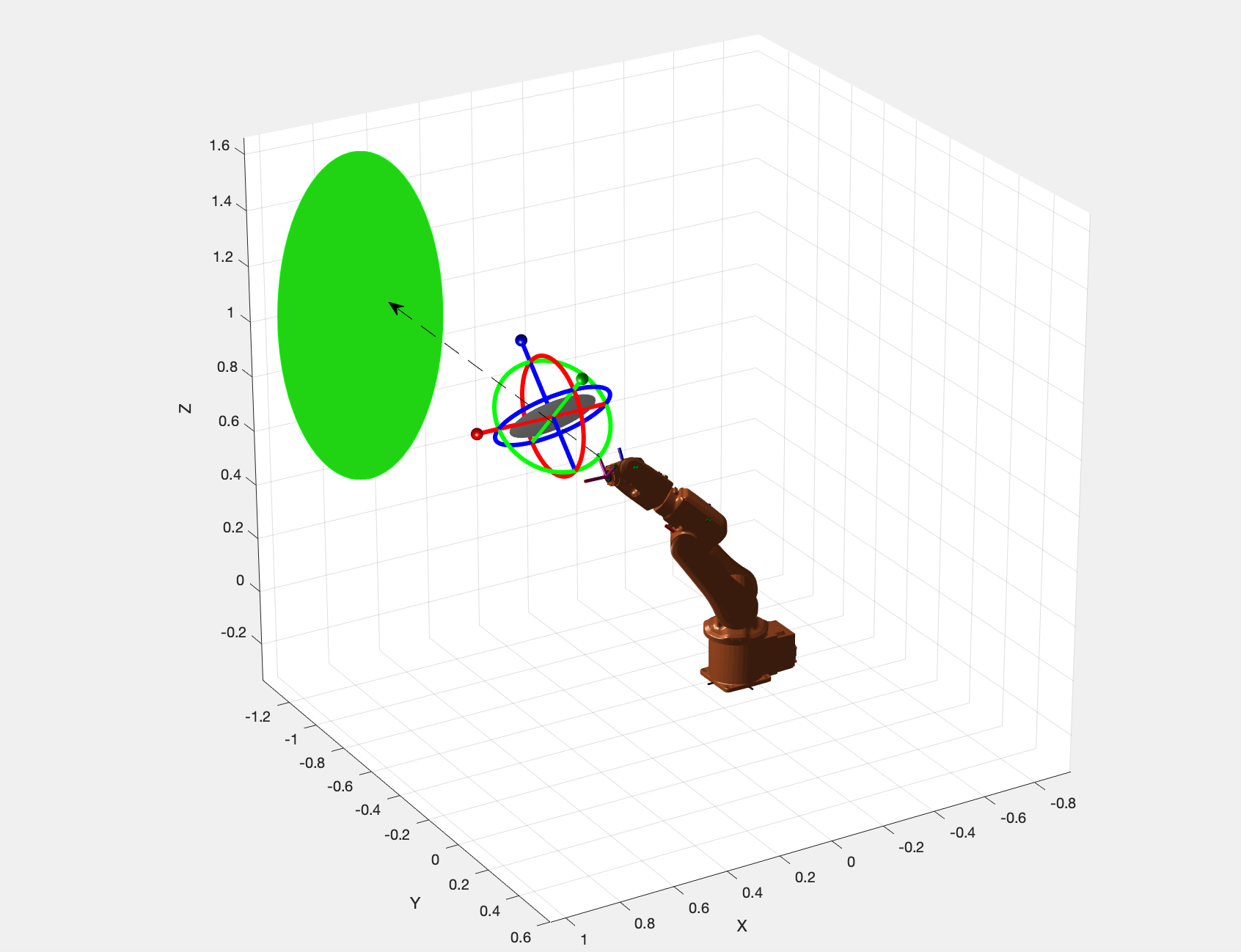} 
    \caption{ABB IRB 120 Manipulator angling hose at plant} 
    \label{fig:robot7}
\end{figure}

\begin{figure}[H]
   \centering	\includegraphics[width=0.75\columnwidth]{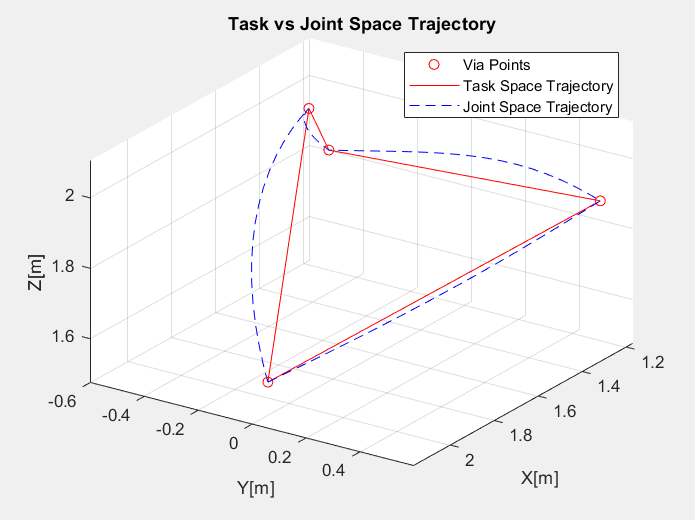} 
    \caption{Joint v. Task Workspace Dynamics} 
    \label{fig:dynamics2}
\end{figure}

    The manipulator is similar, though a miniature toy version, of the 6 DOF ABB IRB 120 industrial arm and its kinematics \cite{Seven:2019} (though not identical given the link length and offset differences.)  However, the workspace of ABB IRB manipulators were simulated since the existing manipulator lacks the torque, robustness, and feedback control to be able to complete the required tasks such as autonomously angling the hose.  Figures \ref{fig:sim0}, \ref{fig:sim1}, and \ref{fig:sim2} simulate the ABB IRB 6620 generating the task workspace dynamics generated in Figures \ref{fig:dynamics} and \ref{fig:dynamics2}.  
Figure \ref{fig:robot7} is an ABB IRB 120 arm simulating aiming its arm in the direction of  a plant. 

\subsection{Joint Trajectory Dynamics}
\begin{figure}[H]
   \centering	
   \includegraphics[width=0.75\columnwidth]{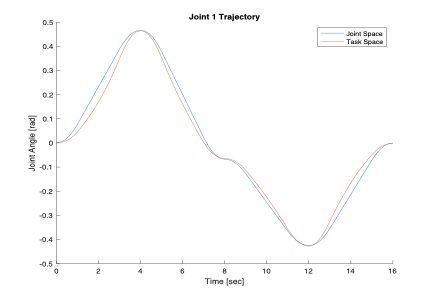} 
    \caption{Joint 1 Trajectory} 
    \label{fig:joint1}
\end{figure}

\begin{figure}[H]
   \centering	
    \includegraphics[width=0.75\columnwidth]{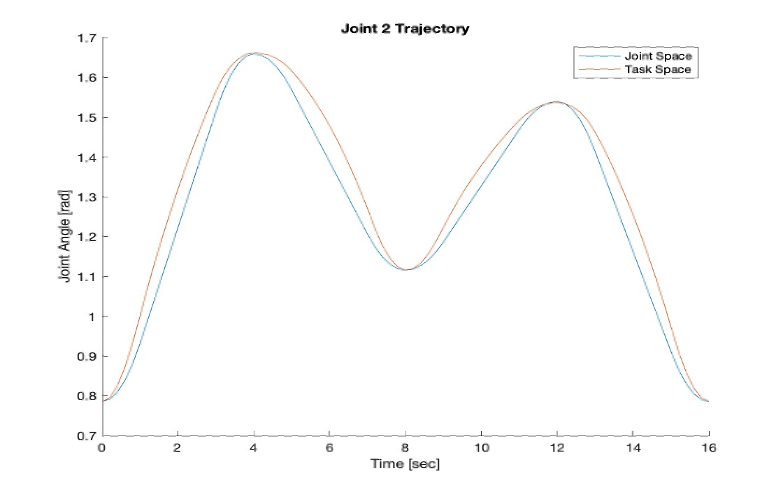} 
    \caption{Joint 2 Trajectory} 
    \label{fig:joint2}
\end{figure}

\begin{figure}[H]
   \centering	
   \includegraphics[width=0.75\columnwidth]{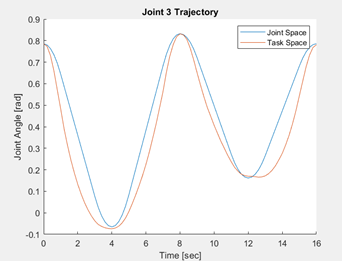} 
    \caption{Joint 3 Trajectory} 
    \label{fig:joint3}
\end{figure}

\begin{figure}[H]
   \centering	\includegraphics[width=0.75\columnwidth]{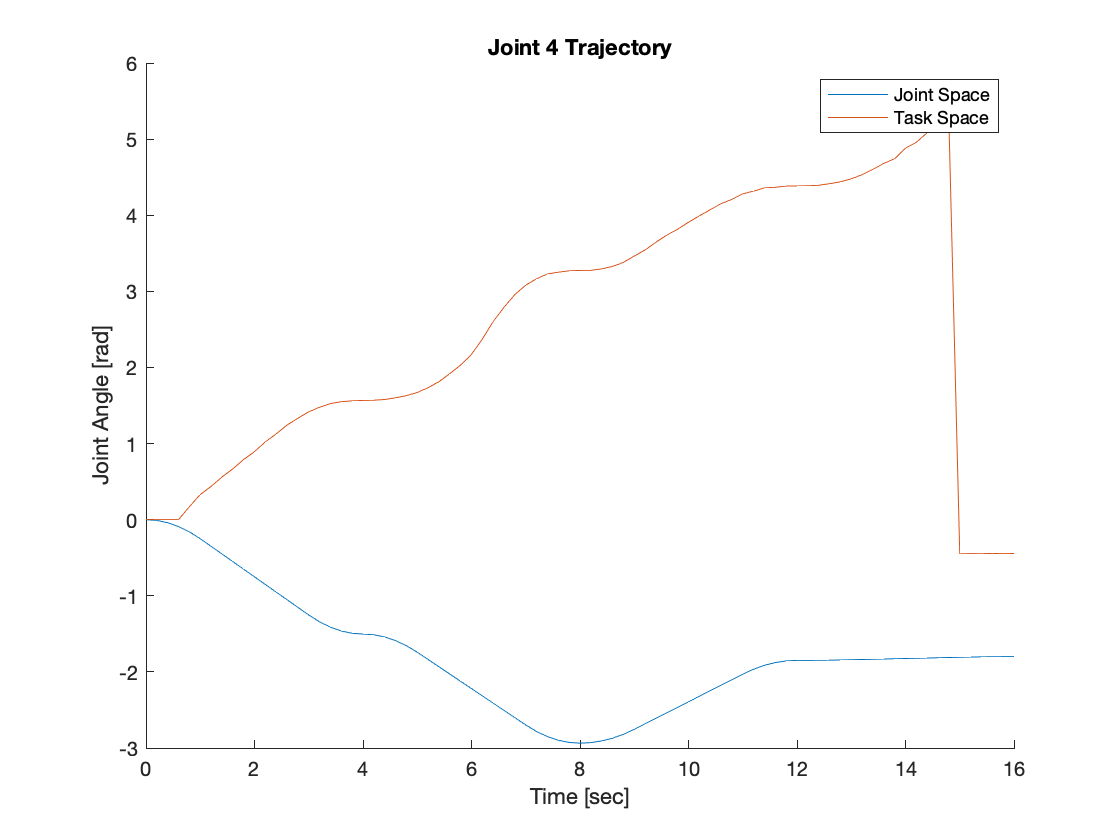} 
    \caption{Joint 4 Trajectory} 
    \label{fig:joint4}
\end{figure}

\begin{figure}[H]
   \centering	
   \includegraphics[width=0.75\columnwidth]{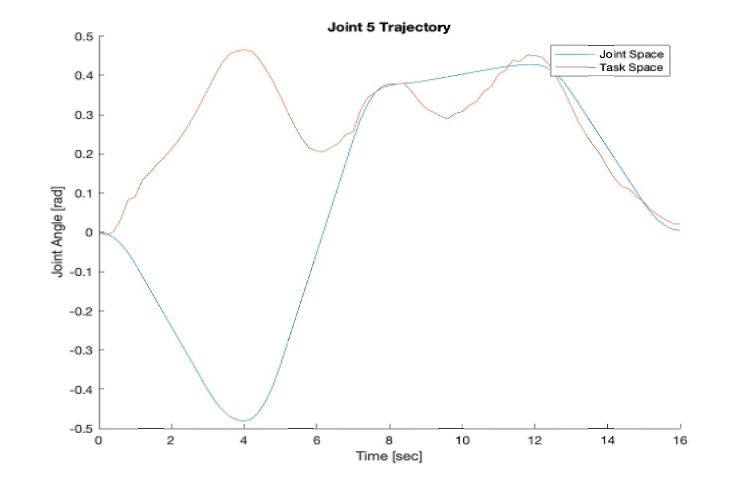} 
    \caption{Joint 5 Trajectory} 
    \label{fig:joint5}
\end{figure}

\begin{figure}[H]
   \centering	
\includegraphics[width=0.75\columnwidth]{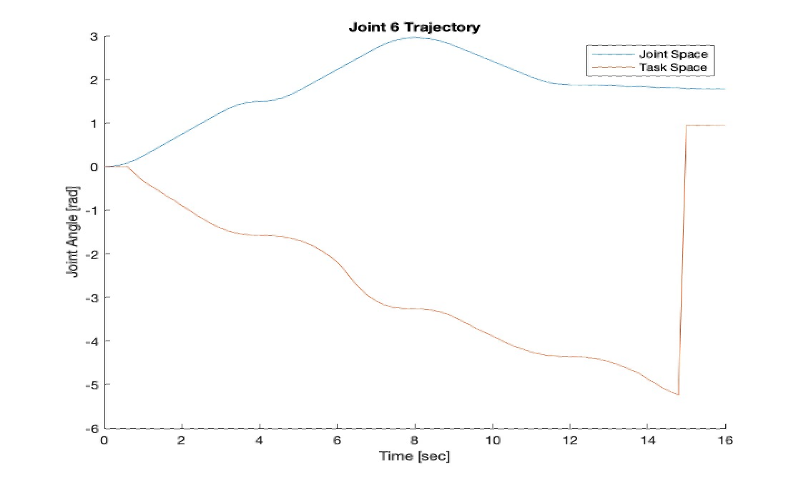} 
    \caption{Joint 6 Trajectory} 
    \label{fig:joint6}
\end{figure}

\begin{figure}[H]
   \centering	
\includegraphics[width=0.75\columnwidth]{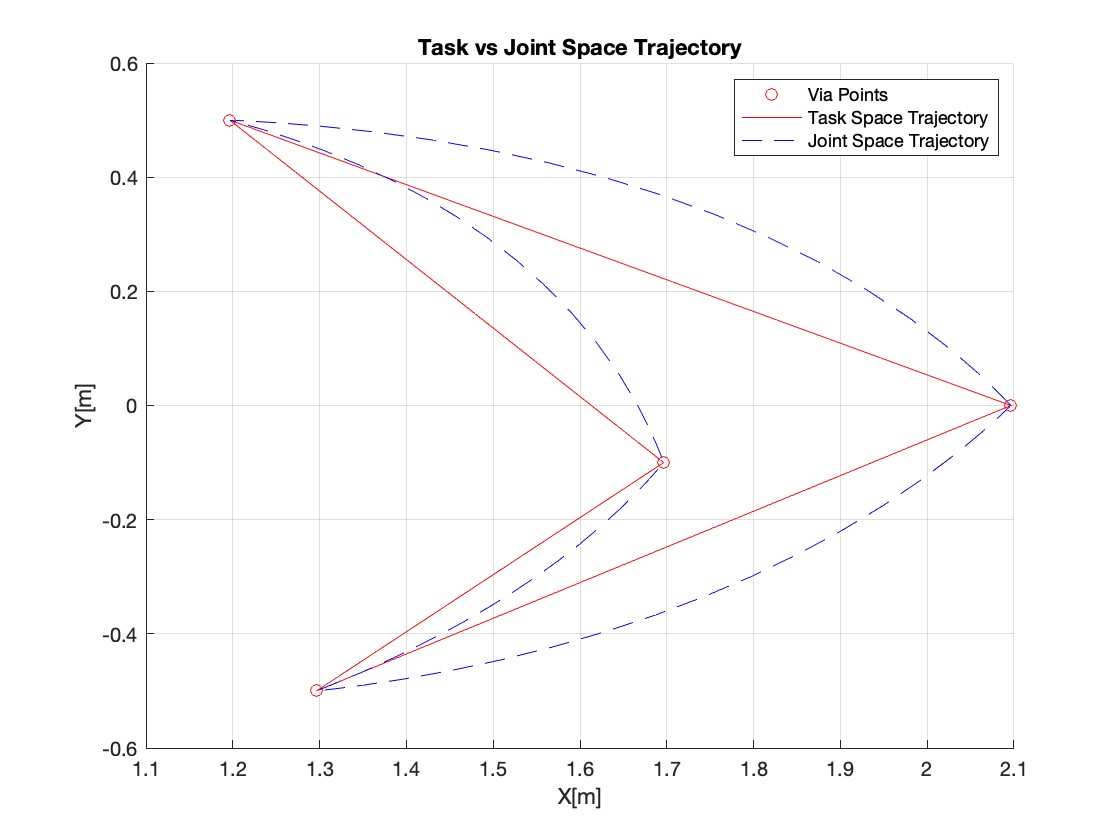} 
    \caption{Joint v. Task Workspace Trajectory} 
    \label{fig:dynamics}
\end{figure}

    Figures \ref{fig:joint1}, \ref{fig:joint2}, \ref{fig:joint3}, \ref{fig:joint4}, \ref{fig:joint5}, and \ref{fig:joint6} show the simulated joint trajectory dynamics of revolute Joint 1 (base) through Joint 6 (wrist), respectively, of the ABB IRB 120 in Matlab that includes torque and feedback-control.  Joint 1 exhibits sinusoidal behavior which one would expect since as the base revolute joint, it rotates and oscillates (through pulse wave signals) the manipulator. 
    Joint v. task workspace simulated dynamics are shown in Figure \ref{fig:dynamics}
     
\subsection{Simulation Dynamics with Chassis}
\begin{figure}[H]
   \centering	\includegraphics[width=0.75\columnwidth]{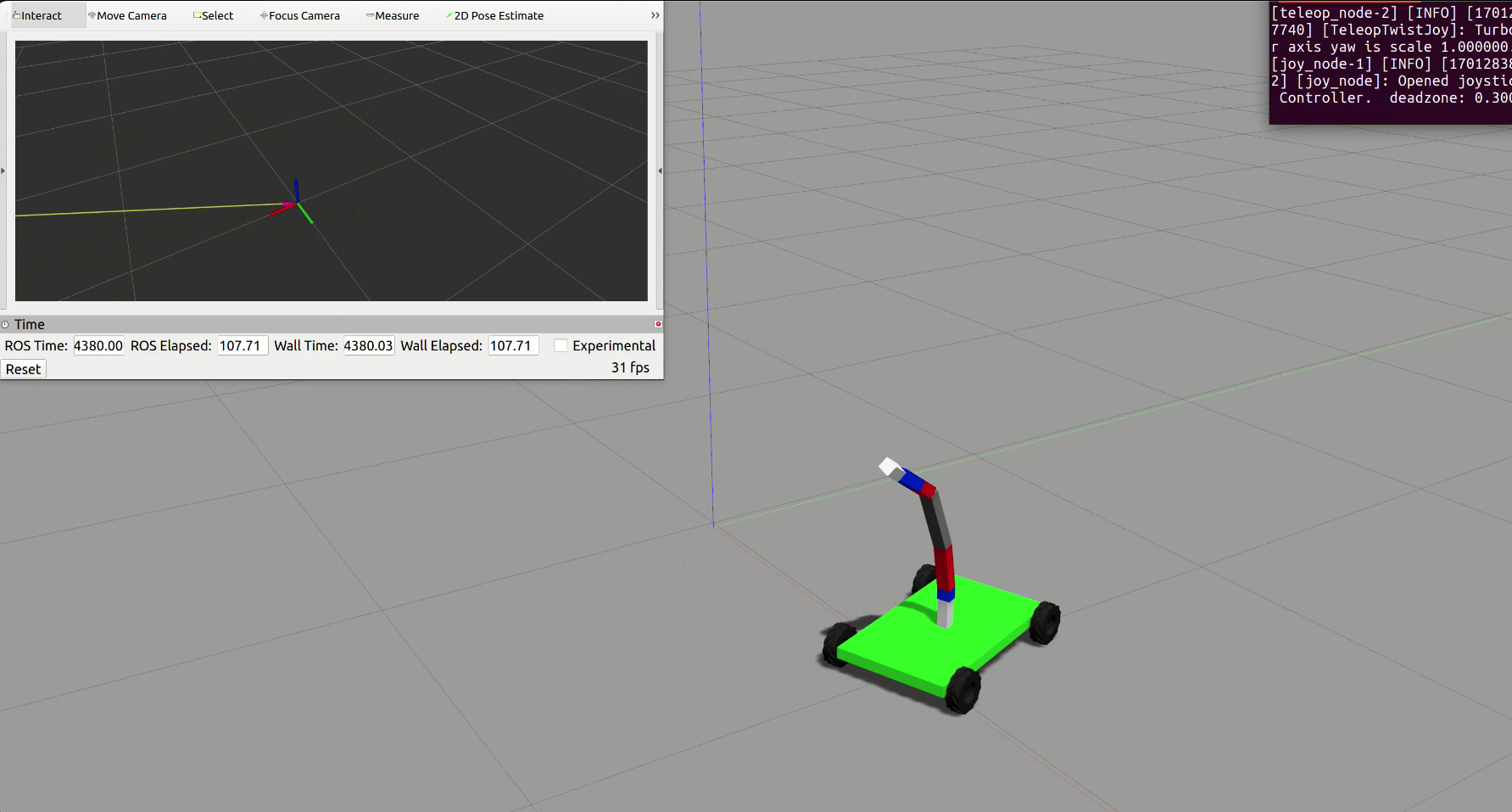} 
    \caption{Gazebo Frame 1} 
    \label{fig:simul0}
\end{figure}

\begin{figure}[H]
   \centering	
\includegraphics[width=0.75\columnwidth]{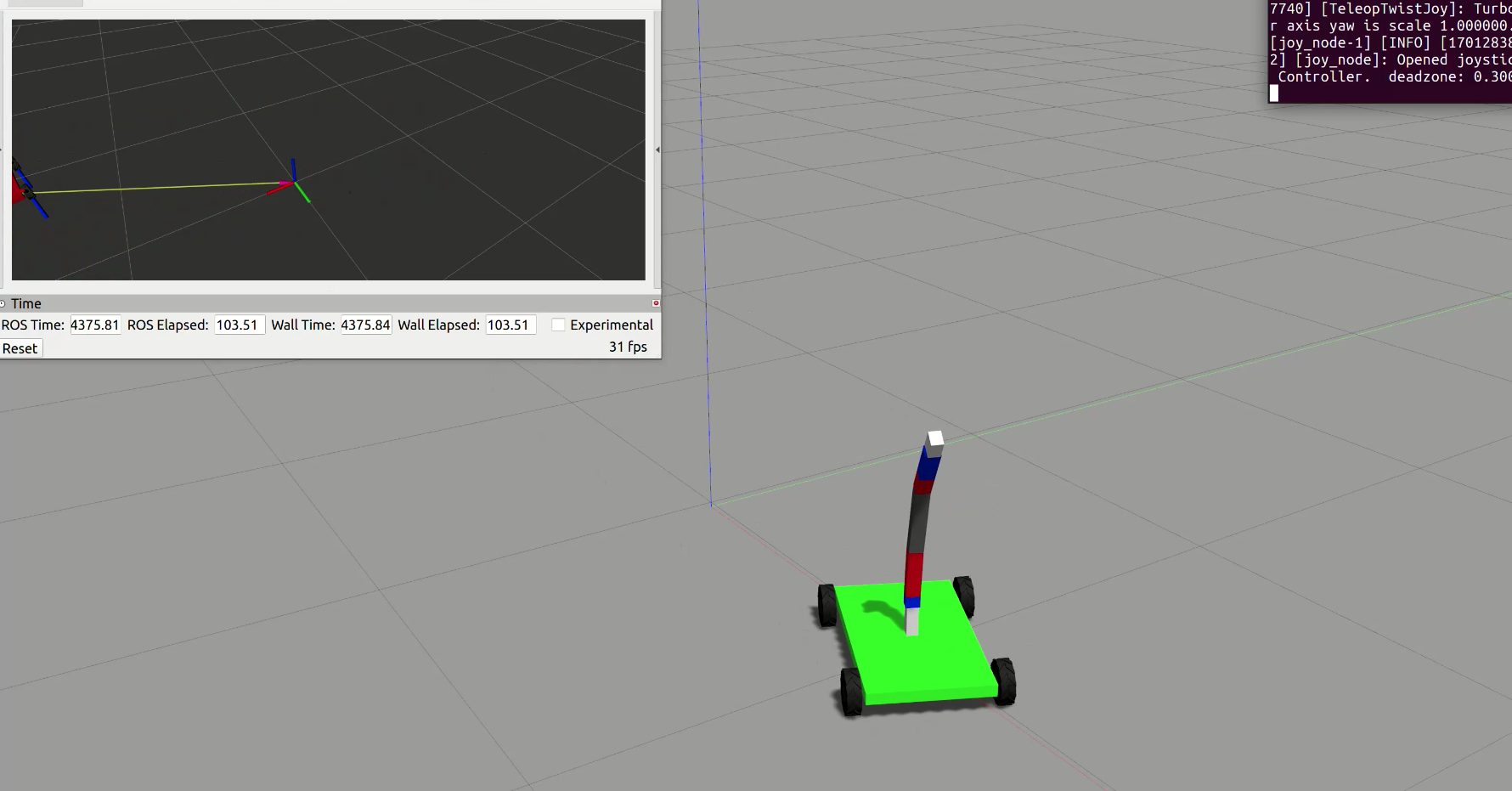} 
    \caption{Gazebo Frame 2} 
    \label{fig:simul1}
\end{figure}

\begin{figure}[H]
   \centering	\includegraphics[width=0.75\columnwidth]{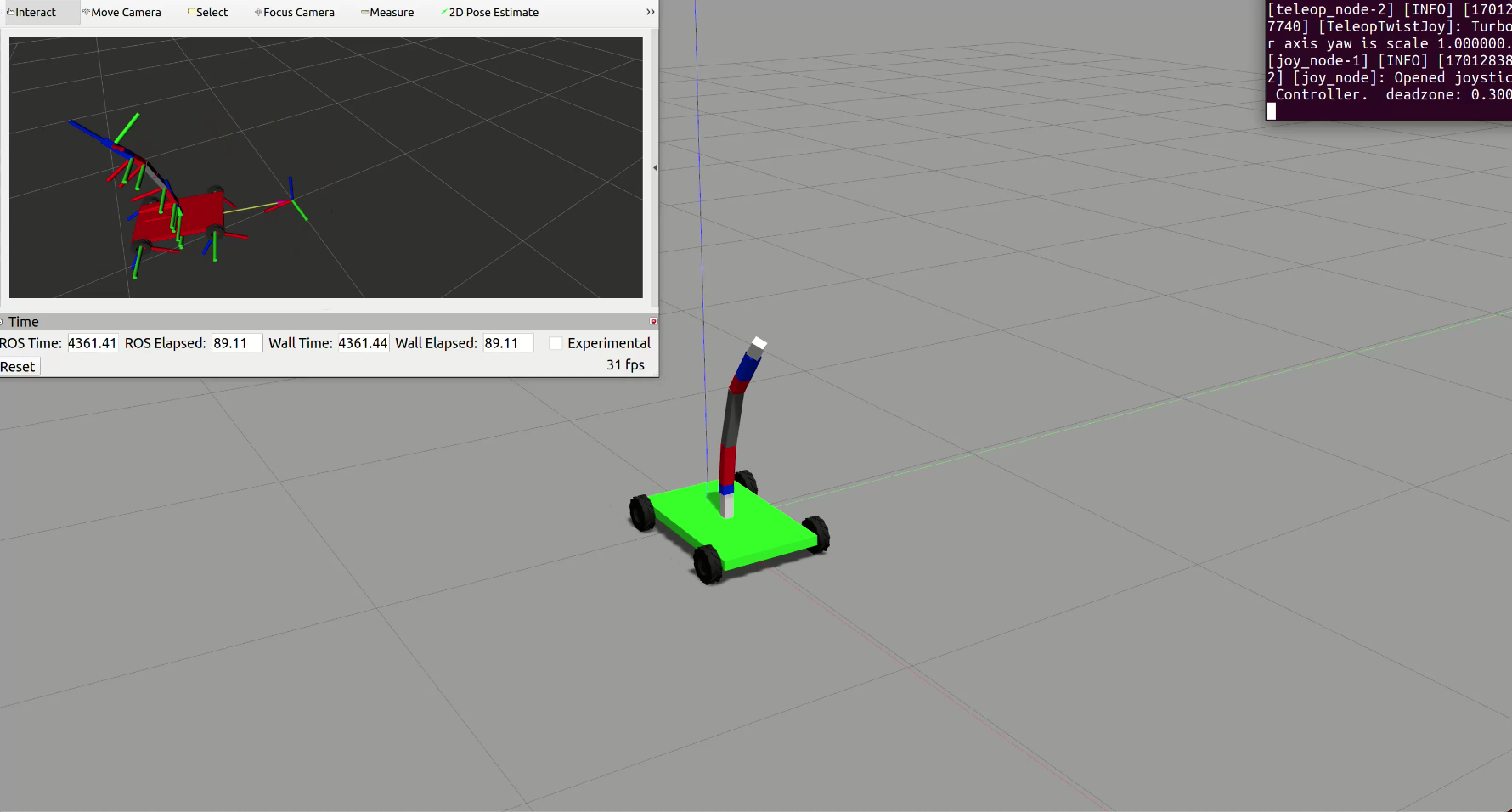} 
    \caption{Gazebo Frame 3} 
    \label{fig:simul2}
\end{figure}

\begin{figure}[H]
   \centering	\includegraphics[width=0.75\columnwidth]{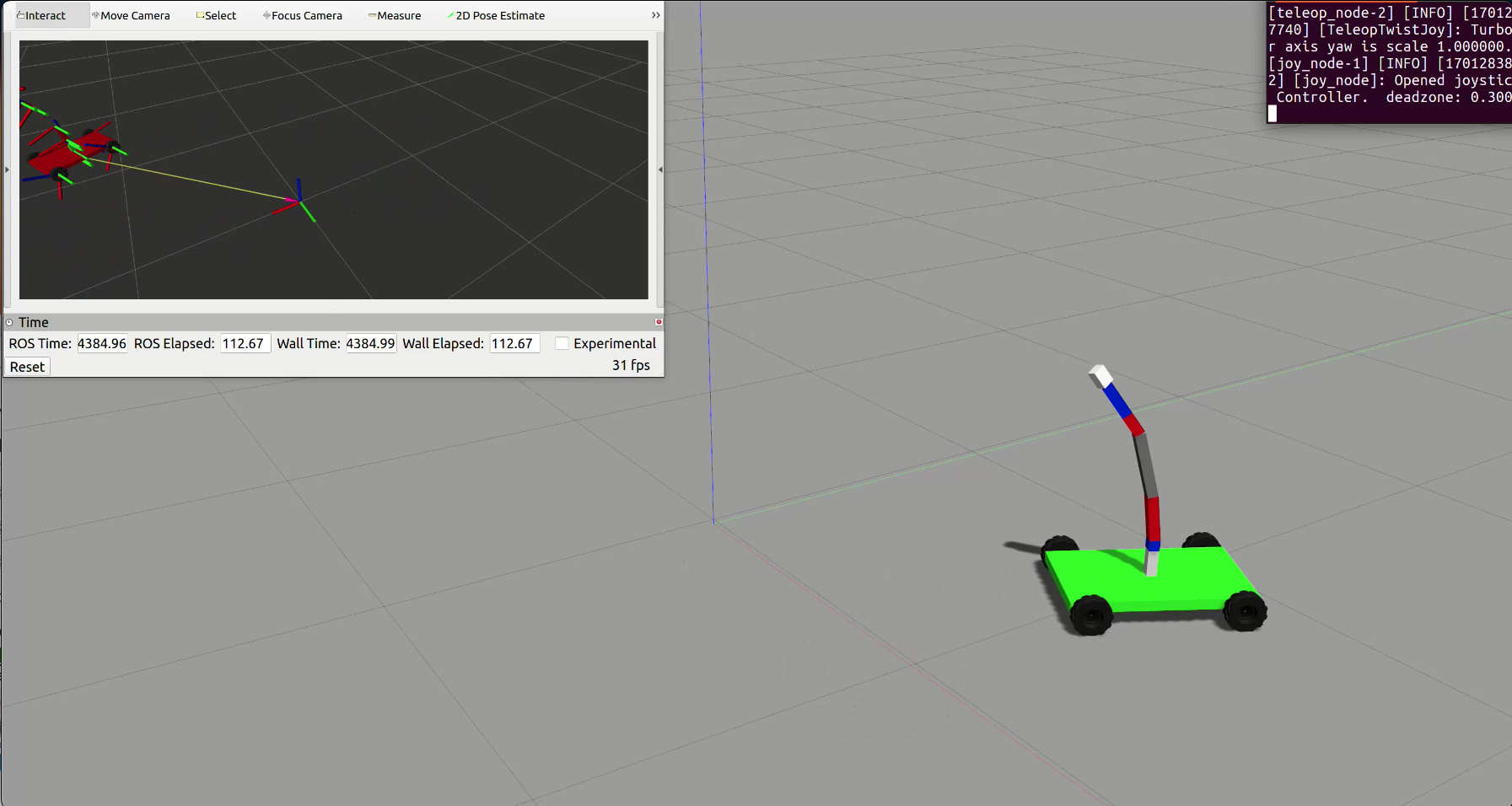} 
    \caption{Gazebo Frame 4} 
    \label{fig:simul3}
\end{figure}

\begin{figure}[H]
   \centering	\includegraphics[width=0.75\columnwidth]{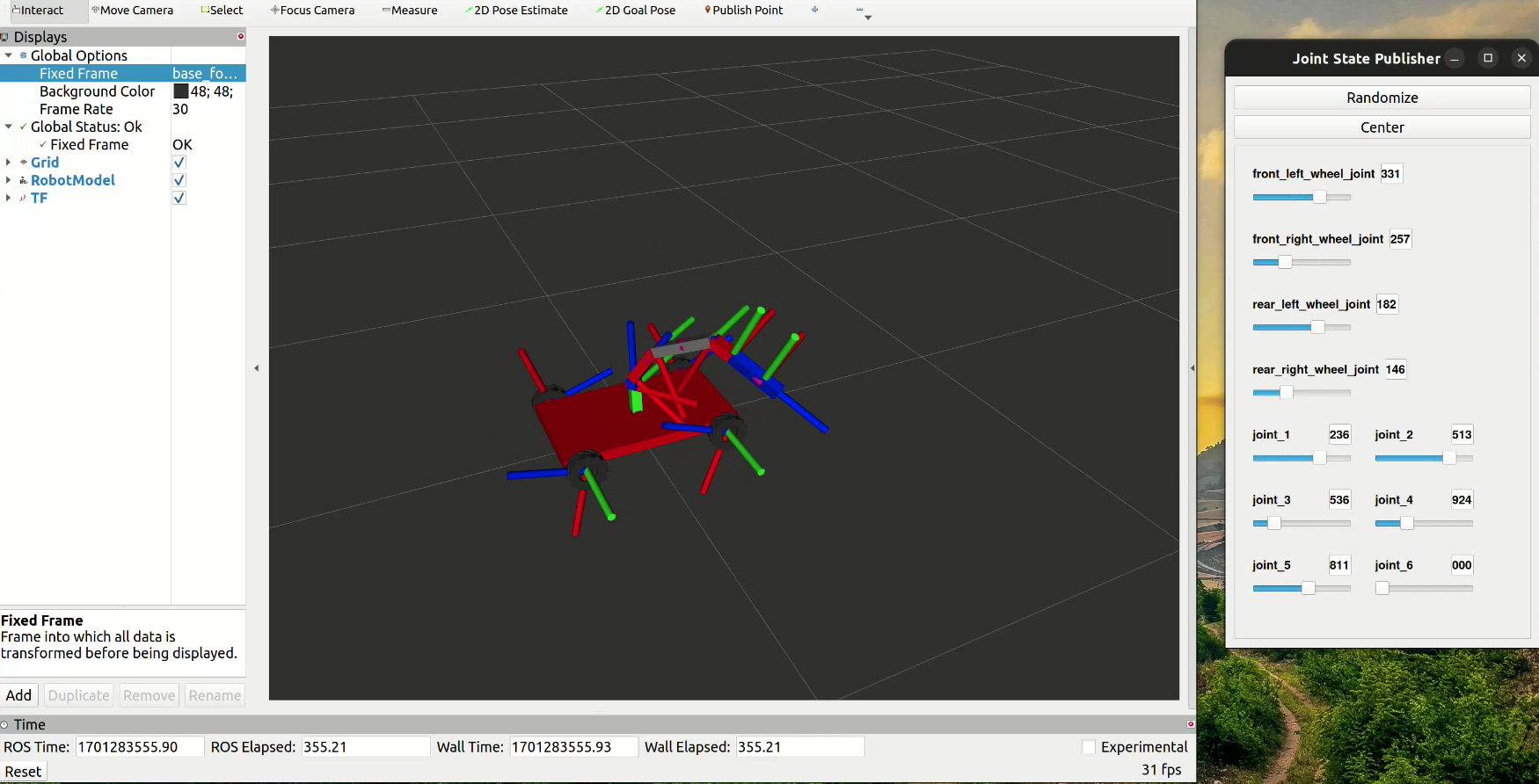} 
    \caption{RViz Frame 1} 
    \label{fig:sim10}
\end{figure}

\begin{figure}[H]
   \centering	
\includegraphics[width=0.75\columnwidth]{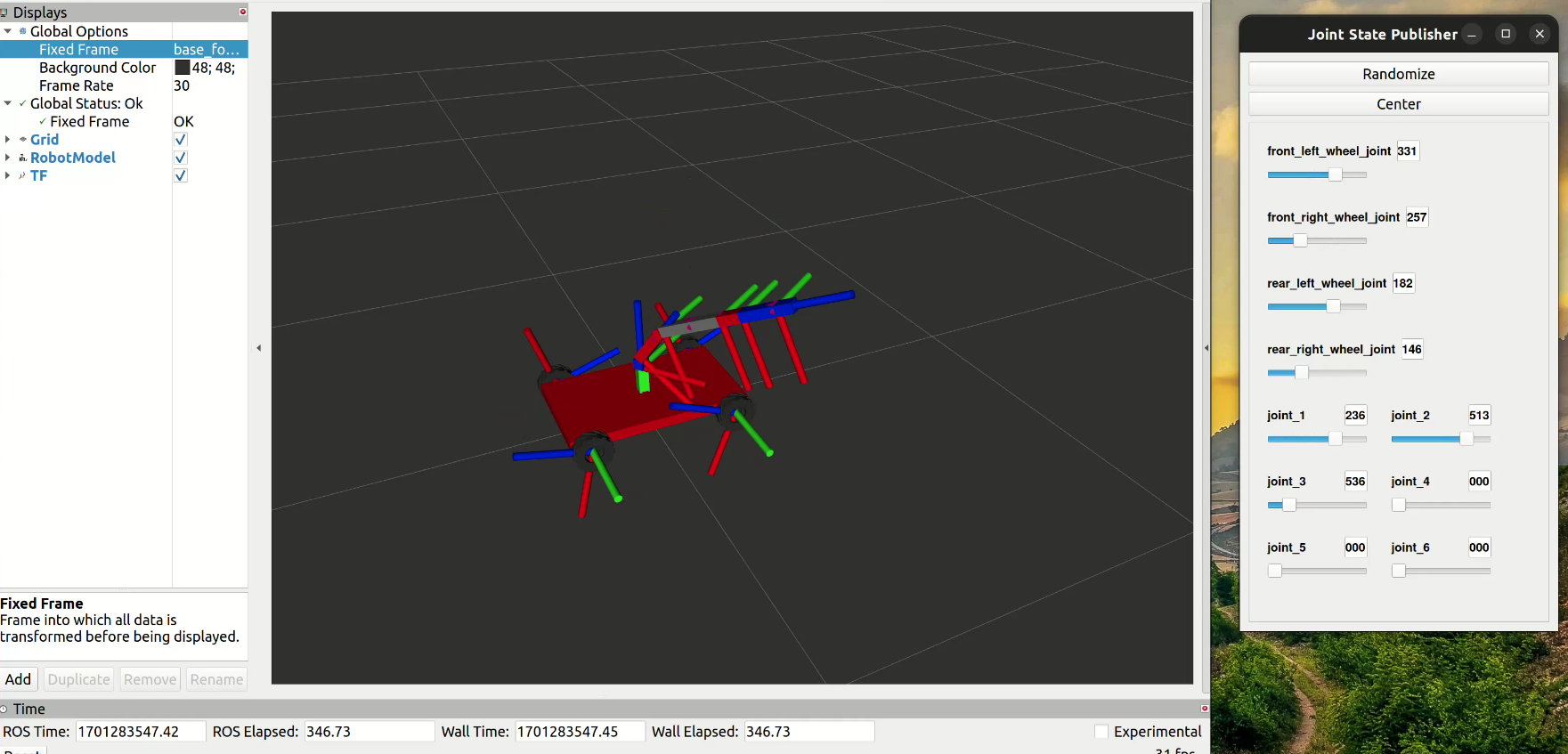} 
    \caption{RViz Frame 2} 
    \label{fig:sim12}
\end{figure}

\begin{figure}[H]
   \centering	\includegraphics[width=0.75\columnwidth]{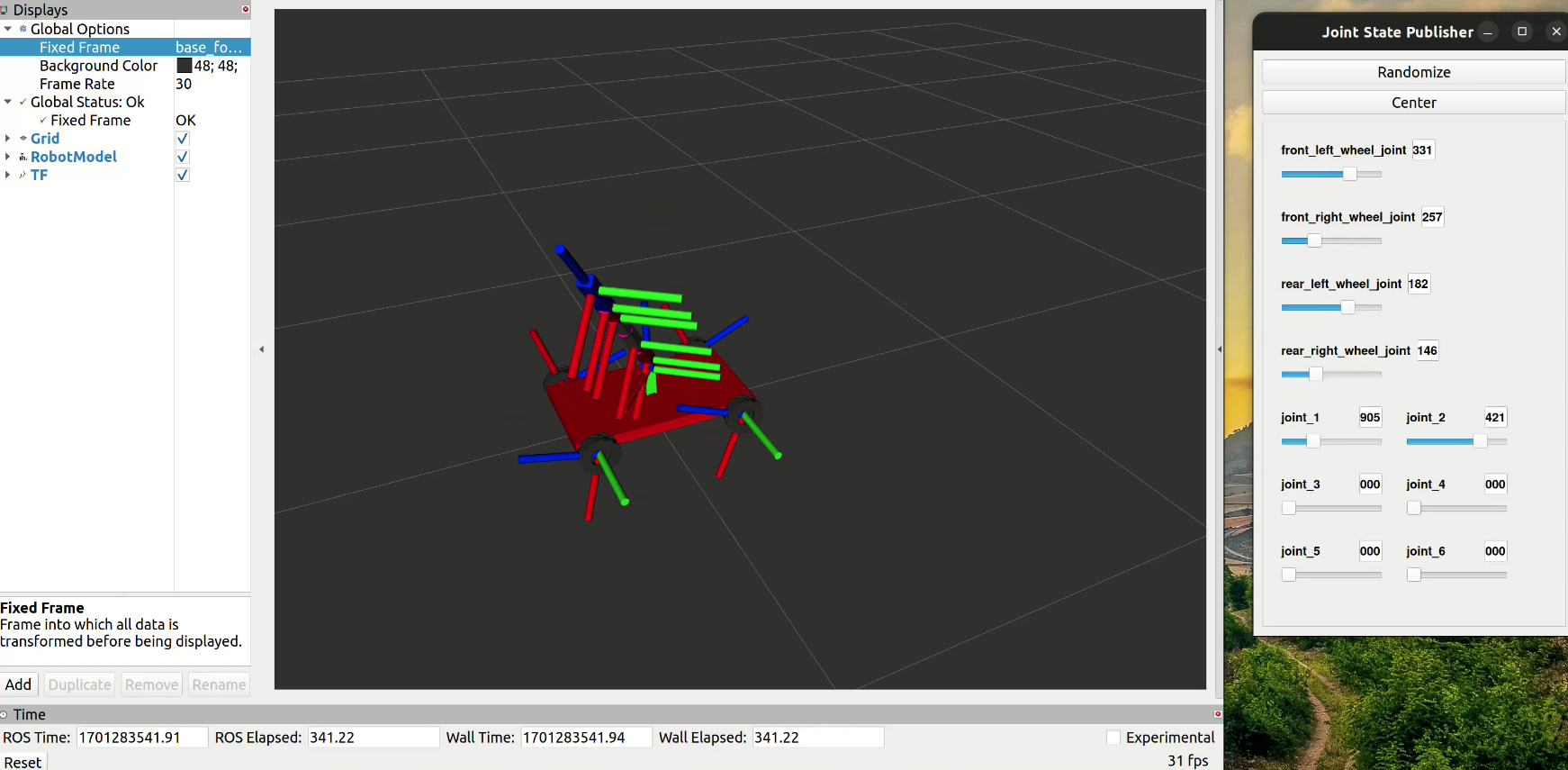} 
    \caption{RViz Frame 3} 
    \label{fig:sim14}
\end{figure}

\begin{figure}[H]
   \centering	\includegraphics[width=0.75\columnwidth]{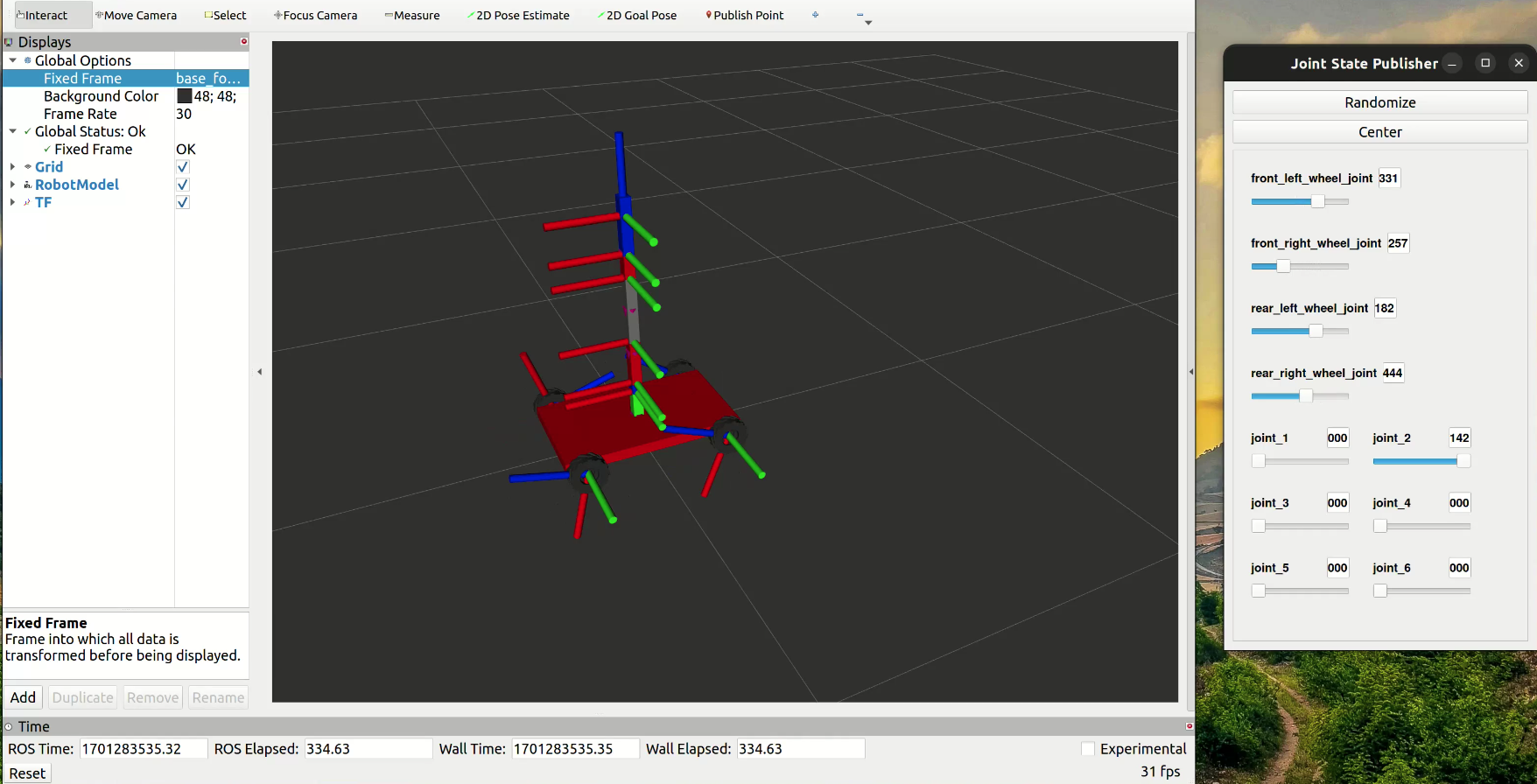} 
    \caption{RViz Frame 4} 
    \label{fig:sim16}
\end{figure}

    \ \ \ Using Gazebo, a robot simulation application for ROS2, we simulate movement of the smart car chassis with the manipulator on it and show various frames in Figures \ref{fig:simul0}, \ref{fig:simul1}, \ref{fig:simul2}, and \ref{fig:simul3}.\footnote{\url{https://classic.gazebosim.org/}}
\ \ \ Using RViz, we simulate the chassis and manipulator forward kinematics and show various frames in Figures \ref{fig:sim10}, \ref{fig:sim12}, \ref{fig:sim14}, and \ref{fig:sim16}
 Given a simulated end-effector pose position (translation) of \textbf{p} = $\begin{bmatrix}
0.04677 \\
0.03679 \\
-0.1343 
\end{bmatrix}$ and a rotation matrix of 
$\mathbf{R} = \begin{bmatrix}
0.3863 & -0.5324 & 0.7532 \\
-0.0466 & 0.8043 & 0.5924 \\
-0.9212 & -0.2639 & 0.2859
\end{bmatrix}$ such that the the base-to-end-effector transformation matrix\\
\\
$\mathbf{T}^{0}_{6} = \begin{bmatrix}
0.3863 & -0.5324 & 0.7532 & 0.04677 \\
-0.0466 & 0.8043 & 0.5924 & 0.03679 \\
-0.9212 & -0.2639 & 0.2859 & -0.1343 \\
0 & 0 & 0 & 1
\end{bmatrix}
$
using joint angles $\mathbf{\theta} = 
\begin{bmatrix} 3.808 & 0.106 & 5.92 &  2.33 &  0.864 &  
1.99
\end{bmatrix}'$, the calculated Jacobian\\ \\$\mathbf{J}
\small
= \begin{bmatrix}
-0.0368 & -0.191 & 0.0922 & -0.0367 & 0 & 0 \\
0.0468 & -0.151 & 0.0725 & -0.0289 & 0 & 0 \\
0 & -0.0595 & 0073 & 0.157 & 0 & 0 \\
0 & -0.0618 & 0.618 & 0.618 & 0.753 & 0.753 \\
0 & 0.786 & -0.786 & -0.786 & 0.592 & 0.592 \\
1 & 0 & 0 & 0 & 0.286 & 0.286 
\end{bmatrix}$, as shown in Appendix \ref{jacobian}.
\section{Experiments}

\ \ \ There were two main tasks (experiments) for the robot to perform to assess its capabilities to carry out tasks.  First, the manipulator was tested to be able hold and then insert a temperature and humidity sensor into the plaint's soil and to obtain measurements.  Second, the manipulator was tested to be able to hold and aim a garden hose to water a plant. Since no experiments could be performed with the manipulator secured onto the alloy chassis, the base of the manipulator had to be tied down by ropes to remain upright to hold the hose.

\begin{figure}[H]
   \centering	\includegraphics[width=0.8\columnwidth]{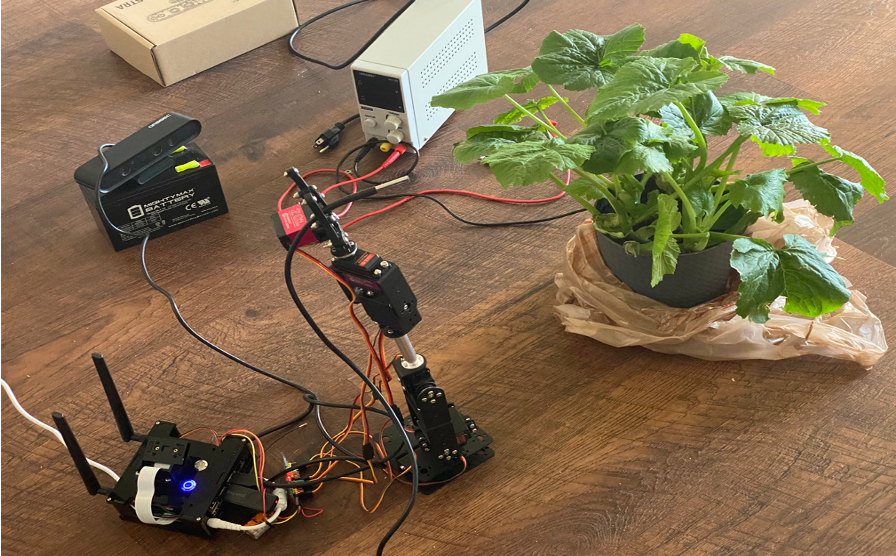} 
    \caption{Manipulator holding sensor}     \label{fig:sensor}
\end{figure}

\begin{figure}[H]
   \centering	\includegraphics[width=0.7\columnwidth]{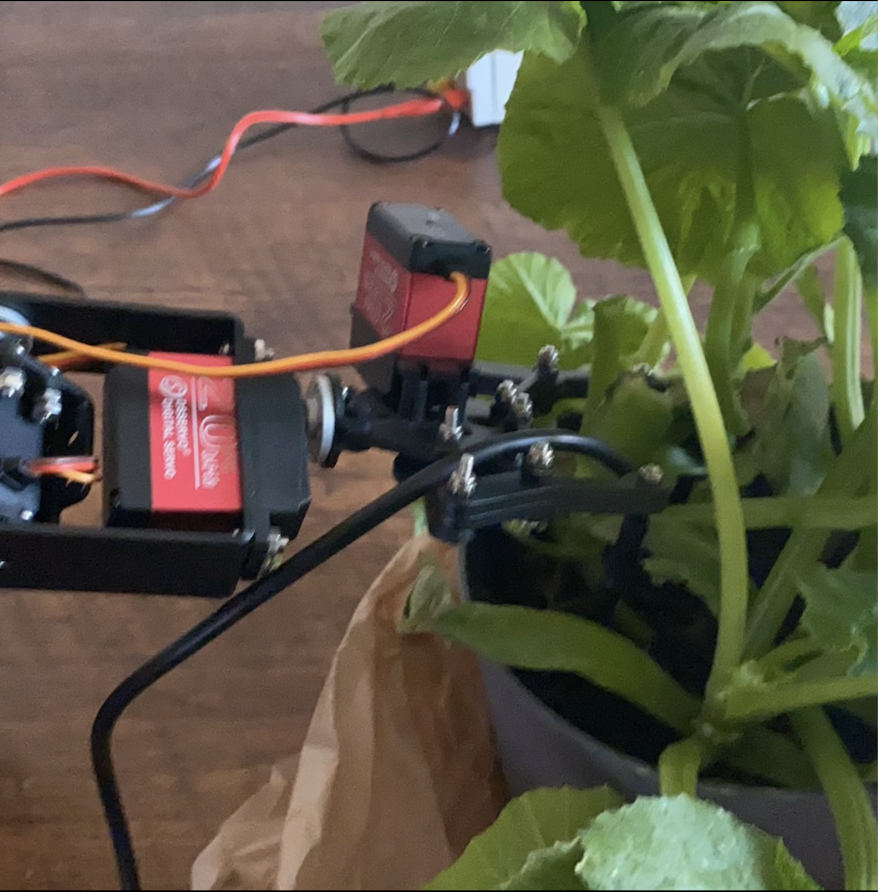} 
    \caption{Manipulator inserting sensor into soil} 
    \label{fig:sensor2}
\end{figure}

\begin{figure}[H]
   \centering	\includegraphics[width=0.7\columnwidth]{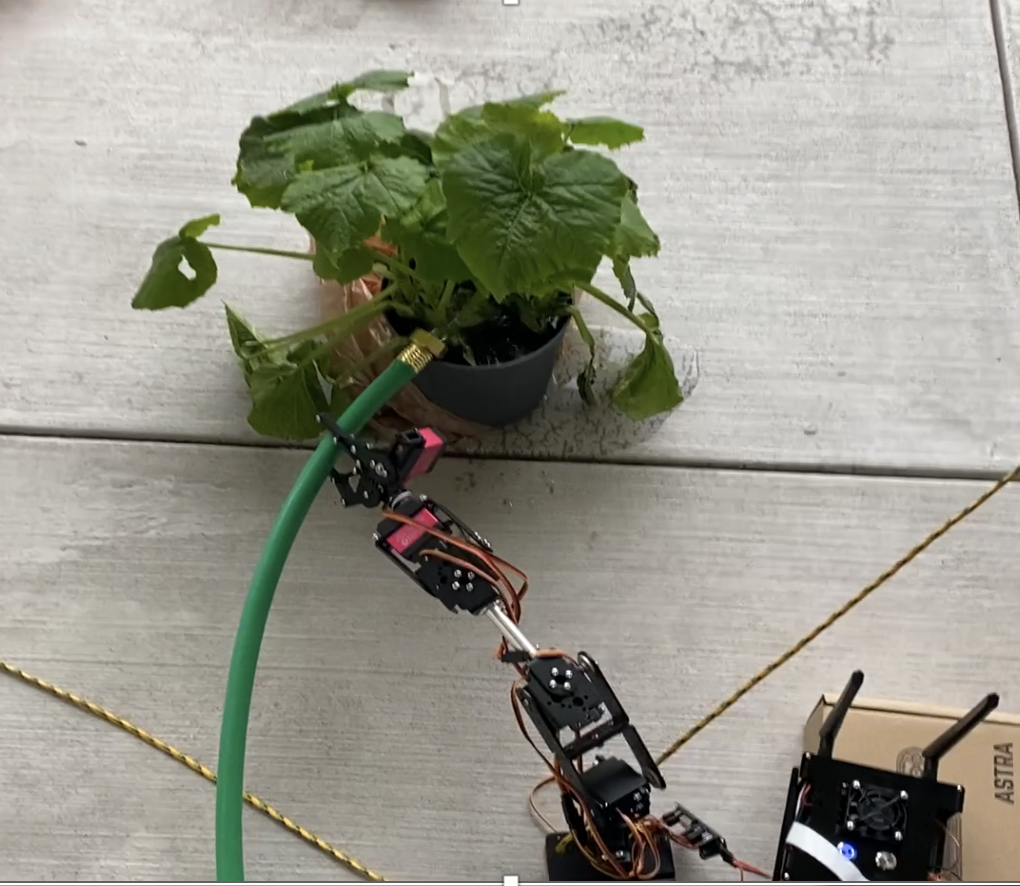} 
    \caption{Manipulator holding and watering plant} 
    \label{fig:hose}
\end{figure}

As shown in Figures \ref{fig:sensor} and \ref{fig:sensor2}, respectively, the manipulator was able to successfully hold and insert the sensor into the soil, but it took several attempts and required placing the plant within the workspace of the manipulator.  Using the Orbbec Astra camera for computer vision for this task on a chassis would have enabled the robot to be able to move to the correct distance from the plant to know how far to hold the hose once variables like the amount of water pressure, type of water spray, and size of the plant are factored into control-feedback calculations.
As shown in Figure \ref{fig:hose}, the manipulator was able to hold the garden hose and water the plant, but was not able to angle it without manual assistance and the hose had to be placed near plant.  This was due to the fact that the manipulator used did not have sufficient torque and was not robust to be able to maneuver the shape, length, and bending of the hose especially as the water pressure increased. For commercial use, a more powerful manipulator is required for these tasks.   
Wiring the various components and keeping the connections and pins secure inside terminals remained a challenge.  While Wago connectors were used to help, given all the cable and jumper wires that crossed one another between connecting microcontrollers to sensors, it is necessary in future work to prevent wire entanglement through future use of single custom-designed printed circuit board (PCB).  
\section{Conclusion}
\ \ \ A novel mobile autonomous plant watering robot was introduced that can overcome limitations of previous robot concepts such as predefined paths for movement and dependence on RFID plant tags as in \cite{Aswani:2012}.  However, there were various challenges in the design of the current robot.  The experiments have shown that a more robust manipulator is required such as a smaller version of the industrial ABB IRR arm that can fit onto a chassis and that has the torque and control-feedback systems to be able to manipulate the hose.  Such a manipulator must be able to adjust the water jet spray levels and properly angle the hose up and down and side to side.  

While this paper discussed the kinematics of water pressure in a hose, the current analysis does not account for such dynamics in the forward and inverse kinematics of the robot, which must be considered in any application of the robot.  For instance, a manipulator that is holding a hose must have proper torque to be able to change the position and angle of the hose to withstand the water pressure, e.g. fire hose exerts enormous water pressure. This remains to be done in future work once a manipulator that has required control-feedback sensors is obtained.  In addition, protection of electronic components from water is essential and therefore this must be considered in future engineering and design of the robot by using a custom-designed closed body shell chassis with the components inside.  Only the computer vision cameras and LiDAR components should be on the outside.

Future work entails incorporating the computer vision plant recognition and objection avoidance/collision experiments using LIDAR once the required chassis is obtained to experiment with the robot's mobility and autonomity.  This will allow experiments when the robot is placed in an actual work environment, in contrast to a virtual one, such a garden or backyard with various obstacles and hazards like swimming pools and rocks.  Additional work also includes programming the robot to keep track of which plants it has watered, the amount of watering each plant needs based on the type of plant and its watering history of the plant which the robot must track.  Finally, the robot must be able to recharge itself from a charging station and be able to autonomously refill its water tank reservoirs when they are low or empty.  With proper engineering redesign, the robot could be manufactured for commercial use and sale.
\appendix
\label{append}
\subsection{URDF Tree of Manipulator}
\label{urdf}
\begin{figure}[H]
   \centering	\includegraphics[width=0.9\columnwidth]{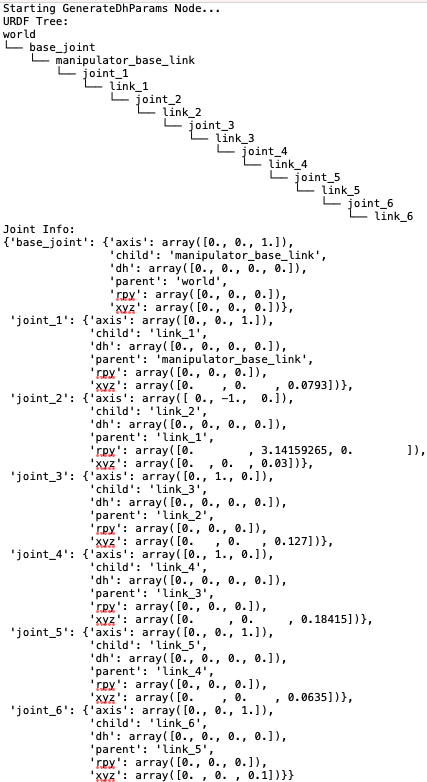} 
    \caption{URDF Tree} 
    \label{fig:urdf}
\end{figure}
\subsection{Elements of Base to End-Effector Transformation Matrix}
\label{elements}
The generation of the matrix elements were calculated in Matlab as shown in the screen shot in Appendix \ref{dh}.
\begin{lstlisting}
n_{x} = cos(theta6)*(sin(theta1)*sin(theta5 + pi/2) - cos(theta5 + pi/2)*(cos(theta4)*(cos(theta1)*sin(theta2)*sin(theta3 - pi/2) - cos(theta1)*cos(theta2)*cos(theta3 - pi/2)) + sin(theta4)*(cos(theta1)*cos(theta2)*sin(theta3 - pi/2) + cos(theta1)*cos(theta3 - pi/2)*sin(theta2)))) + sin(theta6)*(sin(theta5 + pi/2)*(cos(theta4)*(cos(theta1)*sin(theta2)*sin(theta3 - pi/2) - cos(theta1)*cos(theta2)*cos(theta3 - pi/2)) + sin(theta4)*(cos(theta1)*cos(theta2)*sin(theta3 - pi/2) + cos(theta1)*cos(theta3 - pi/2)*sin(theta2))) + cos(theta5 + pi/2)*sin(theta1))

o_{x} = cos(theta6)*(sin(theta5 + pi/2)*(cos(theta4)*(cos(theta1)*sin(theta2)*sin(theta3 - pi/2) - cos(theta1)*cos(theta2)*cos(theta3 - pi/2)) + sin(theta4)*(cos(theta1)*cos(theta2)*sin(theta3 - pi/2) + cos(theta1)*cos(theta3 - pi/2)*sin(theta2))) + cos(theta5 + pi/2)*sin(theta1)) - sin(theta6)*(sin(theta1)*sin(theta5 + pi/2) - cos(theta5 + pi/2)*(cos(theta4)*(cos(theta1)*sin(theta2)*sin(theta3 - pi/2) - cos(theta1)*cos(theta2)*cos(theta3 - pi/2)) + sin(theta4)*(cos(theta1)*cos(theta2)*sin(theta3 - pi/2) + cos(theta1)*cos(theta3 - pi/2)*sin(theta2)))) 

a_{x} = cos(theta4)*(cos(theta1)*cos(theta2)*sin(theta3 - pi/2) + cos(theta1)*cos(theta3 - pi/2)*sin(theta2)) - sin(theta4)*(cos(theta1)*sin(theta2)*sin(theta3 - pi/2) - cos(theta1)*cos(theta2)*cos(theta3 - pi/2)) 

p_{x} = 3*cos(theta1))/100 + (389*sin(theta1))/1250 + (327*sin(theta1)*sin(theta5 + pi/2))/2000 - (327*cos(theta5 + pi/2)*(cos(theta4)*(cos(theta1)*sin(theta2)*sin(theta3 - pi/2) - cos(theta1)*cos(theta2)*cos(theta3 - pi/2)) + sin(theta4)*(cos(theta1)*cos(theta2)*sin(theta3 - pi/2) + cos(theta1)*cos(theta3 - pi/2)*sin(theta2))))/2000 + 793/10000

n_{y} = - cos(theta6)*(cos(theta1)*sin(theta5 + pi/2) - cos(theta5 + pi/2)*(cos(theta4)*(cos(theta2)*cos(theta3 - pi/2)*sin(theta1) - sin(theta1)*sin(theta2)*sin(theta3 - pi/2)) - sin(theta4)*(cos(theta2)*sin(theta1)*sin(theta3 - pi/2) + cos(theta3 - pi/2)*sin(theta1)*sin(theta2)))) - sin(theta6)*(cos(theta1)*cos(theta5 + pi/2) + sin(theta5 + pi/2)*(cos(theta4)*(cos(theta2)*cos(theta3 - pi/2)*sin(theta1) - sin(theta1)*sin(theta2)*sin(theta3 - pi/2)) - sin(theta4)*(cos(theta2)*sin(theta1)*sin(theta3 - pi/2) + cos(theta3 - pi/2)*sin(theta1)*sin(theta2))))

o_{y} = sin(theta6)*(cos(theta1)*sin(theta5 + pi/2) - cos(theta5 + pi/2)*(cos(theta4)*(cos(theta2)*cos(theta3 - pi/2)*sin(theta1) - sin(theta1)*sin(theta2)*sin(theta3 - pi/2)) - sin(theta4)*(cos(theta2)*sin(theta1)*sin(theta3 - pi/2) + cos(theta3 - pi/2)*sin(theta1)*sin(theta2)))) - cos(theta6)*(cos(theta1)*cos(theta5 + pi/2) + sin(theta5 + pi/2)*(cos(theta4)*(cos(theta2)*cos(theta3 - pi/2)*sin(theta1) - sin(theta1)*sin(theta2)*sin(theta3 - pi/2)) - sin(theta4)*(cos(theta2)*sin(theta1)*sin(theta3 - pi/2) + cos(theta3 - pi/2)*sin(theta1)*sin(theta2)))) 

a_{y} = cos(theta4)*(cos(theta2)*sin(theta1)*sin(theta3 - pi/2) + cos(theta3 - pi/2)*sin(theta1)*sin(theta2)) + sin(theta4)*(cos(theta2)*cos(theta3 - pi/2)*sin(theta1) - sin(theta1)*sin(theta2)*sin(theta3 - pi/2))     

p_{y} = 3*sin(theta1))/100 - (389*cos(theta1))/1250 - (327*cos(theta1)*sin(theta5 + pi/2))/2000 + (327*cos(theta5 + pi/2)*(cos(theta4)*(cos(theta2)*cos(theta3 - pi/2)*sin(theta1) - sin(theta1)*sin(theta2)*sin(theta3 - pi/2)) - sin(theta4)*(cos(theta2)*sin(theta1)*sin(theta3 - pi/2) + cos(theta3 - pi/2)*sin(theta1)*sin(theta2))))/2000

n_{z} = cos(theta6)*cos(theta5 + pi/2)*(sin(theta4)*(cos(theta2)*cos(theta3 - pi/2) - sin(theta2)*sin(theta3 - pi/2)) + cos(theta4)*(cos(theta2)*sin(theta3 - pi/2) + cos(theta3 - pi/2)*sin(theta2))) - sin(theta6)*sin(theta5 + pi/2)*(sin(theta4)*(cos(theta2)*cos(theta3 - pi/2) - sin(theta2)*sin(theta3 - pi/2)) + cos(theta4)*(cos(theta2)*sin(theta3 - pi/2) + cos(theta3 - pi/2)*sin(theta2)))

o_{z} = -cos(theta6)*sin(theta5 + pi/2)*(sin(theta4)*(cos(theta2)*cos(theta3 - pi/2) - sin(theta2)*sin(theta3 - pi/2)) + cos(theta4)*(cos(theta2)*sin(theta3 - pi/2) + cos(theta3 - pi/2)*sin(theta2))) - cos(theta5 + pi/2)*sin(theta6)*(sin(theta4)*(cos(theta2)*cos(theta3 - pi/2) - sin(theta2)*sin(theta3 - pi/2)) + cos(theta4)*(cos(theta2)*sin(theta3 - pi/2) + cos(theta3 - pi/2)*sin(theta2)))

a_{z} = sin(theta4)*(cos(theta2)*sin(theta3 - pi/2) + cos(theta3 - pi/2)*sin(theta2)) - cos(theta4)*(cos(theta2)*cos(theta3 - pi/2) - sin(theta2)*sin(theta3 - pi/2))

p_{z} = 327*cos(theta5 + pi/2)*(sin(theta4)*(cos(theta2)*cos(theta3 - pi/2) - sin(theta2)*sin(theta3 - pi/2)) + cos(theta4)*(cos(theta2)*sin(theta3 - pi/2) + cos(theta3 - pi/2)*sin(theta2))))/2000
\end{lstlisting}
\subsection{DH Table Analysis}
\label{dh}
\begin{figure}[H]
   \centering	\includegraphics[width=0.9\columnwidth]{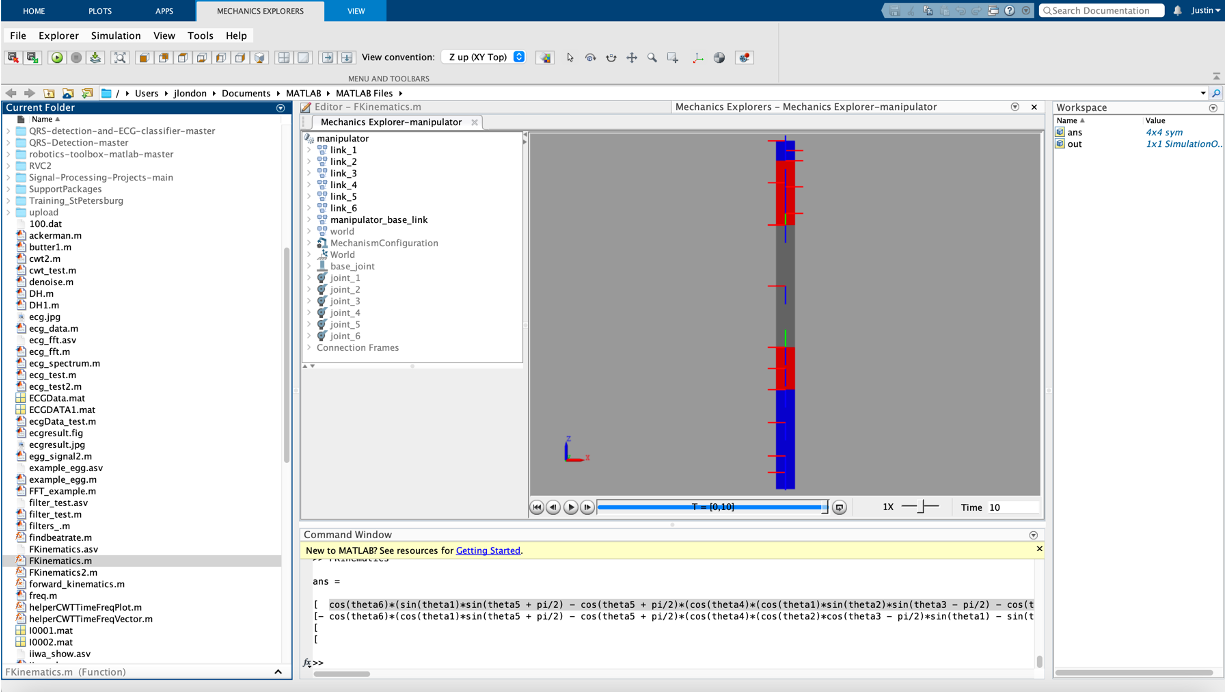} 
    \caption{Matlab Coordinate Analysis}
    \label{fig:links}   
\end{figure}
\begin{figure}[H]
   \centering	\includegraphics[width=0.75\columnwidth]{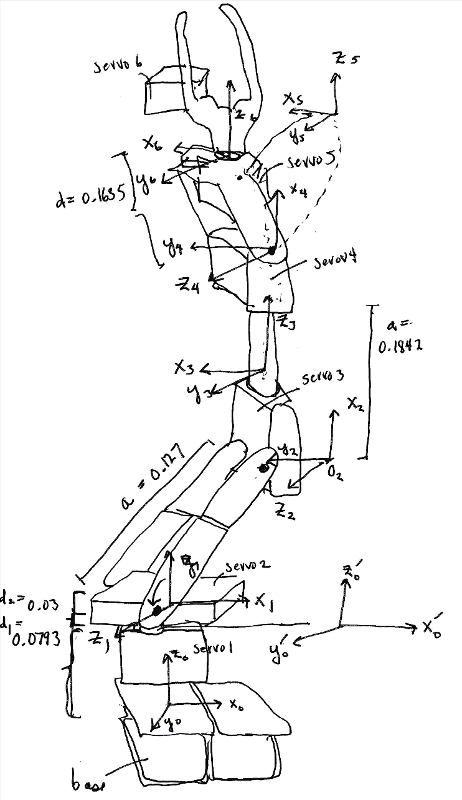} 
    \caption{DH Coordinates by Hand}
    \label{fig:matlab10}   
\end{figure}
\begin{figure}[H]
   \centering	\includegraphics[width=0.8\columnwidth]{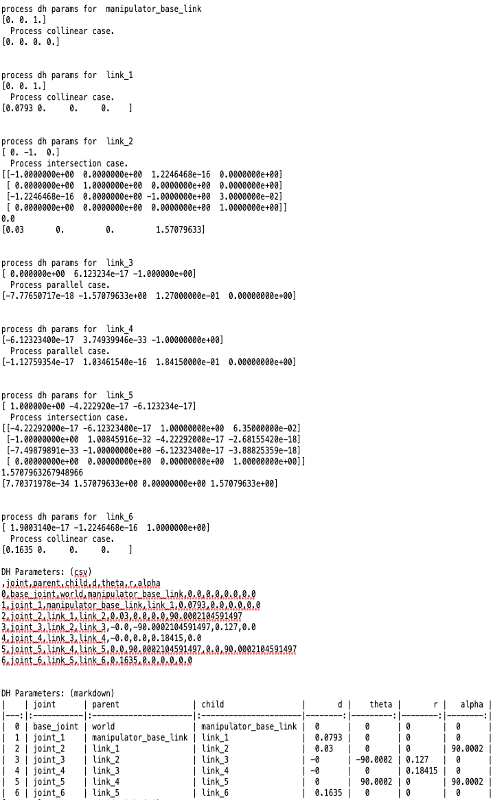}
    \label{fig:dh}   
    \caption{DH Computations}
\end{figure}
\subsection{Architecture}
\label{architecture}
\begin{figure}[H]
   \centering	\includegraphics[width=1.2\columnwidth]{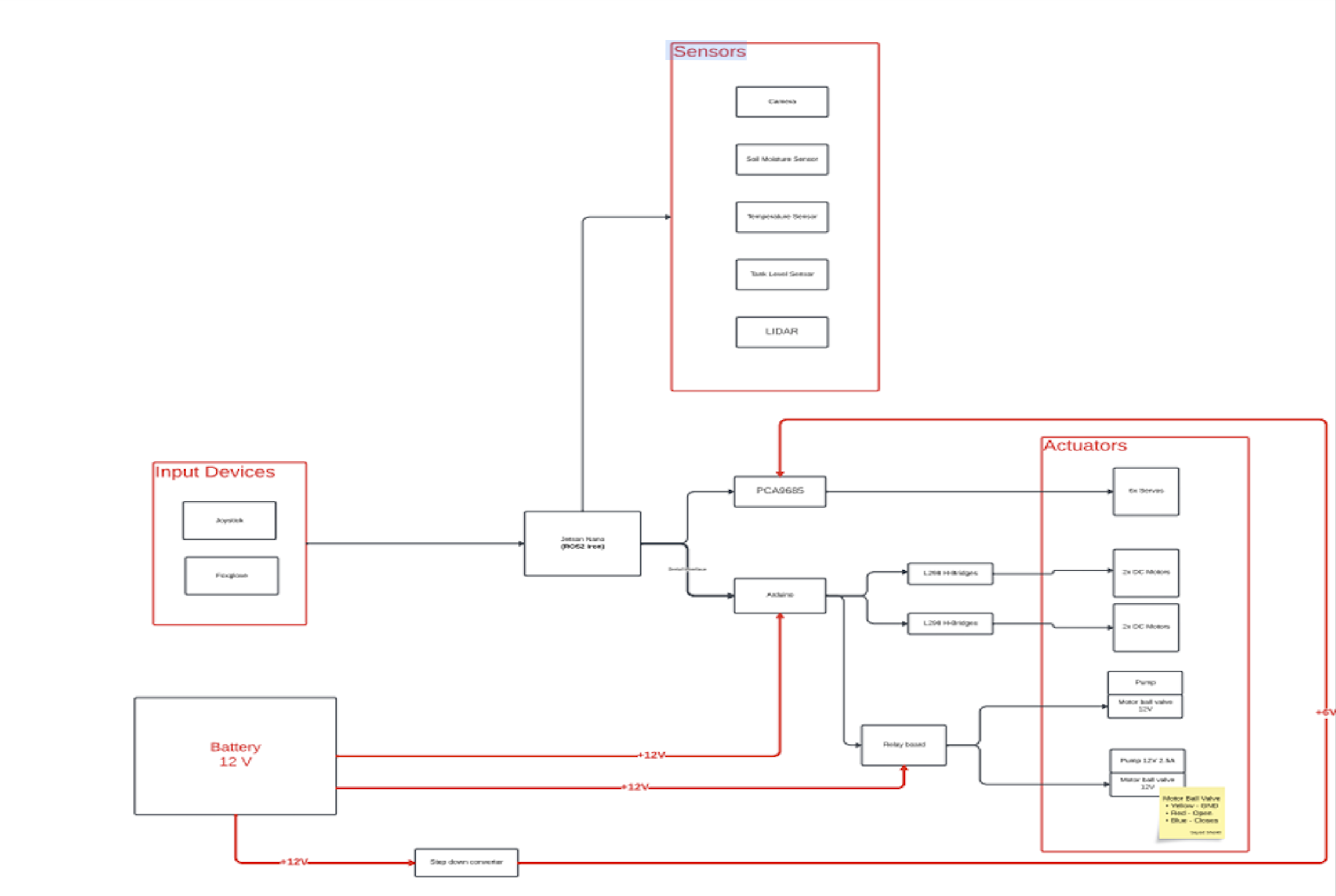} 
    \caption{Robot Architecture}
    \label{fig:architecture}   
\end{figure}
\subsection{Jacobian}
\label{jacobian}
\begin{figure}[H]
   \centering	\includegraphics[width=0.9\columnwidth]{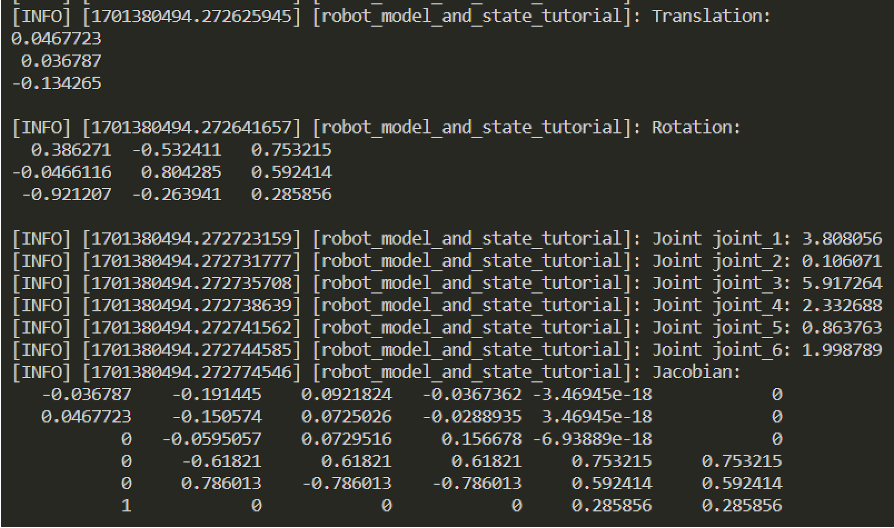}
    \label{fig:jacobian}   
\end{figure}

\printbibliography
\end{document}